\pdfoutput=1
\documentclass[11pt]{article}

\usepackage[preprint]{acl}

\usepackage{times}
\usepackage{latexsym}
\usepackage[T1]{fontenc}
\usepackage[utf8]{inputenc}
\usepackage{microtype}
\usepackage{graphicx}
\usepackage{amsmath}
\usepackage{amssymb}
\usepackage{amsthm}
\newtheorem{proposition}{Proposition}
\usepackage{booktabs}
\usepackage{tabularx}
\usepackage{multirow}
\usepackage{makecell}
\usepackage{float}
\usepackage{placeins}
\usepackage[normalem]{ulem}
\usepackage{xspace}
\usepackage{url}
\usepackage[most]{tcolorbox}
\usepackage{etoc}

\usepackage{xcolor}
\usepackage{colortbl}
\newcommand{\heat}[1]{\pgfmathsetmacro{\heatlvl}{max(0,min(45,150*(1.25-#1)))}\edef\heatcell{\noexpand\cellcolor{green!\heatlvl}}\heatcell#1}

\newcommand{\taubench}{\mbox{$\tau^3$-bench}\xspace}
\newcommand{\ours}{BiPA\xspace}

\newtcolorbox{promptbox}[1]{
  enhanced,
  breakable,
  colback=blue!5!white,
  colframe=blue!75!black,
  coltitle=white,
  fonttitle=\bfseries,
  title=#1,
  arc=2mm,
  boxrule=1pt,
  width=\linewidth
}
\newtcolorbox{responsebox}[1]{
  enhanced,
  breakable,
  colback=green!5!white,
  colframe=green!55!black,
  coltitle=white,
  fonttitle=\bfseries,
  title=#1,
  arc=2mm,
  boxrule=1pt,
  width=\linewidth
}
\tcbset{claimbox/.style={
  enhanced,
  boxrule=0pt,
  leftrule=2pt,
  sharp corners,
  left=4pt, right=4pt, top=2pt, bottom=2pt,
  before skip=5pt, after skip=5pt,
  fontupper=\small
}}
\newcounter{finding}
\newtcolorbox{finding}[1][]{
  claimbox,
  colback=blue!4!white,
  colframe=blue!60!black,
  before upper={\refstepcounter{finding}\textbf{Finding~\thefinding\if\relax\detokenize{#1}\relax\else\ (#1)\fi.}\ }
}
\newcounter{principle}
\newtcolorbox{principle}{
  claimbox,
  colback=orange!6!white,
  colframe=orange!80!black,
  before upper={\refstepcounter{principle}\textbf{Principle~\theprinciple.}\ }
}

\title{Does Your Generalist Skill Optimizer Really Incentivize Agentic Tasks?\\Rethinking Skill Optimization via Bipartite Attribution}

\author{%
  \textbf{Jiawei Zhou\textsuperscript{1,2} \quad Bo Peng\textsuperscript{1,3} \quad Yuxuan Huang\textsuperscript{1,3} \quad Ming Zhang\textsuperscript{1}} \\
  \textbf{Youbang Sun\textsuperscript{1,5}\thanks{Corresponding authors.} \quad Xiang Wang\textsuperscript{1,4} \quad Chaochao Lu\textsuperscript{1}\footnotemark[1]} \\[3pt]
  {\textsuperscript{1}Shanghai AI Laboratory} \\
  {\textsuperscript{2}Fudan University \quad \textsuperscript{3}Shanghai Jiao Tong University} \\
  {\textsuperscript{4}University of Science and Technology of China \quad \textsuperscript{5}Tsinghua University}
}
\hypersetup{
  pdftitle={Does Your Generalist Skill Optimizer Really Incentivize Agentic Tasks? Rethinking Skill Optimization via Bipartite Attribution},
  pdfauthor={Jiawei Zhou, Bo Peng, Yuxuan Huang, Ming Zhang, Youbang Sun, Xiang Wang, Chaochao Lu}
}

\begin{document}
\etocdepthtag.toc{main}
\maketitle

\begin{abstract}
{Skill optimizers adapt an agent to long-horizon tasks, such as repairing a terminal environment, by rewriting its skill, a natural-language procedure in its context, and keeping the edits that raise success on a small dev set. On such tasks success is noisy, so a gain on a small dev set does not show that an edit helps, and a success score does not show what the edit costs. We introduce \ours{}, which replays an optimizer's acceptance rule on candidates identical to its seed to measure what selection alone reports, and records each episode as a bipartite graph of turns and the text they read to trace extra input tokens to the behavior behind them. Across three optimizers, three agentic benchmarks and two models, the learned skills succeed about as often as their hand-written seed, 0.1 points higher on average. Their dev gains vanish on test: keeping the best of several candidates on at most 30 tasks reports gains this large even when none changed. They read about 16\% more input tokens: the optimizers almost always add rules, half to re-check work and some to act one call at a time, and the agent follows them as extra turns, each re-sending the whole conversation. Four principles follow: accept an edit only if its gain exceeds what selection alone reports; charge it for the turns it adds, not for its length; let the editor delete and merge rules; show it what earlier rules cost. Code is available at {\url{https://github.com/valleysprings/BiPA.git}}.}

\end{abstract}

\section{Introduction}
\label{sec:intro}

\begin{figure*}[t]
\centering
\includegraphics[width=\textwidth]{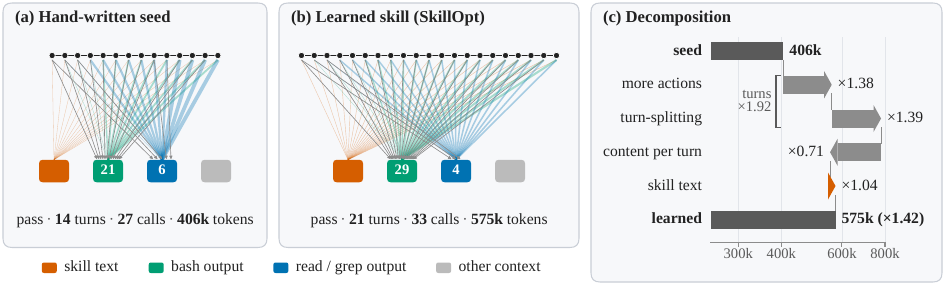}
\caption{\textbf{On one AutomationBench task, SkillOpt's skill passes like the seed but reads 1.42 times its input tokens, through extra turns.} (a) Seed, (b) learned skill; box numbers: tool calls. (c) The ratio as a product of four factors.}
\label{fig:teaser}
\end{figure*}

{A \emph{harness}, such as OpenCode or SWE-agent, is the agentic scaffold that runs a language model as an agent: it gives the model its prompt and tools and runs its loop until the task is done \citep{opencode,swe_agent}. Built for software engineering, these harnesses now run agents on other long-horizon tasks, such as repairing a terminal environment, serving a customer or running a business workflow. An agent is adapted to a family of such tasks with a \emph{skill}, a natural-language procedure that the harness places in the model's context at the start of every task. Skill optimizers automate this, starting from a \emph{seed} skill \citep{gepa,skillopt,skilladam}: an \emph{editor} model reads the \emph{trajectories}, the logs of the agent's runs on training tasks, and turns their successes and failures into rules, and a \emph{gate} keeps an edit if the agent with the edited skill solves more tasks of a small development set, the dev set. These optimizers report gains, mostly on short, cheap tasks such as question answering, and controlled studies find the gains sparse and fragile \citep{liu2026rethinking,wang2026rethinkingharness}.}

Long-horizon tasks expose three gaps. First, selection is noisy. Long tasks keep dev sets small and success is stochastic, so one run can rank two skills either way \citep{miller2024errorbars,kapoor2024agentsmatter}, and keeping the best of several comparisons reports a gain even when nothing changed, a winner's curse \citep{smith_winkler_2006}: a reported gain does not show that a skill helps. Second, success hides cost. Each turn, one request to the model, re-sends the whole history, so extra turns raise the input tokens read over a task, and a success score does not show them. Third, rules go uncredited. The editor writes several rules per edit and the gate scores the whole edit with one number, so no rule is tested alone and it stays unknown which rules change the agent.

We propose \ours{} (Bipartite Attribution; Figure~\ref{fig:teaser}), which measures these gaps from the agent's logs without changing the optimizer. For the first gap, it replays the optimizer's acceptance rule on candidates identical to the seed, so any gain the rule reports comes from selection alone (Section~\ref{sec:replay}). For the second, it records each attempt at a task as a bipartite graph between the agent's turns and the text each turn reads (Section~\ref{sec:graph}) and splits a rise in input tokens into more tool calls, the same calls over more turns, more text per turn and longer skill text (Sections~\ref{sec:elasticity} and~\ref{sec:split}). For the third, we sort each added rule into one of eight categories by what it asks the agent to do and compare the categories with the behavior the graph records (Section~\ref{sec:analysis}).

{We apply \ours{} to GEPA, SkillOpt and SkillAdam on Terminal-Bench 2.1 (TB2), \taubench{} and AutomationBench (AutoBench) with DeepSeek-V4.1-Flash and GPT-6-Luna, and test transfer by running some of DeepSeek-V4.1-Flash's skills under GPT-6-Luna or on another benchmark. All start from one seed, a generic hand-written procedure from a public repository that asks for a fresh, complete check before any claim of completion \citep{superpowers}.}

{Learned skills succeed about as often as the seed: over all sixteen, test success is 0.1 points higher. The dev gains the optimizers report do not hold on test: the gate keeps the best of several candidates on a dev set of at most 30 tasks, where unchanged candidates would report gains of the same size. The skills gain about as many tasks as they lose, and read about 16\% more input tokens than the seed, over more turns. The edits grow the skill: 91 of 97 scored edits lengthen it, about half of the rules they add tell the agent to re-check its work and none to read less, and the agent carries these rules out as extra turns, each re-sending the whole conversation. The skills do not carry over to another model or benchmark, and most of their cost goes with them.}

Our contributions are:
\begin{itemize}\itemsep1pt\parsep0pt
    \item We formulate \ours{}, which dissects skill optimization without changing the optimizer: a selection replay gives the dev gain that selection alone reports, and a bipartite graph of turns and context traces each extra input token to the behavior behind it (Sections~\ref{sec:replay} to~\ref{sec:split}).
    \item Based on \ours{}, we show why learned skills do not pay off: selection alone can produce their dev gains, and their rules add turns that each re-send the whole conversation, a cost that, unlike the gain, travels with the skill to other models and benchmarks.
    \item Based on this mechanism, we propose four design principles for skill optimizers (Section~\ref{sec:implications}): accept an edit only if its gain exceeds what selection alone reports; charge it for the turns it adds, not for its length; let the editor delete and merge rules; and show it what its earlier rules cost.
\end{itemize}

\section{Related Work}
\label{sec:related}

\subsection{Text Optimization for Agents}

\paragraph{Prompt and skill optimization.} {Prompt optimizers let a language model edit scored prompts or follow textual feedback or evolutionary search \citep{ape,opro,textgrad,evoprompt}; GEPA has a model reflect on the logs of its runs and draws the skill to edit from the skills that are best on at least one dev task \citep{gepa}. Skill optimizers apply this loop to the procedures an agent harness loads: SkillOpt accepts edits of limited size that raise dev success \citep{skillopt}, SkillAdam adapts its edit size \citep{skilladam}, and {EvoSkill keeps a skill when it raises held-out success \citep{evoskill}; Trace2Skill and experience memories distill skills or playbooks from trajectories \citep{trace2skill,agent_workflow_memory,expel,ace_context_engineering}; all learn from task outcomes.}}

\paragraph{Cost-aware optimization.} {A few optimizers make cost a goal: {CAPO penalizes prompt length and MO-CAPO the input and output tokens of inference} \citep{capo,mo_capo}, SkillMOO searches for skills that are best on pass rate and cost together \citep{skillmoo}, SkillReducer and SkillZip Pro compress skill text \citep{skillreducer,skillzip}, and cost-aware rewriting finds that shorter skills can cost more \citep{xing2026skillremember}.}

\subsection{Agent and Harness Evaluation}

\paragraph{Evaluating optimizer gains.} {Controlled studies find that skill evolution is sparse and filtered by validation, with validation and test {sometimes favoring} different feedback \citep{liu2026rethinking}; {that on Terminal-Bench 2.1 harness evolution gains little on held-out tasks and does not beat simply running the agent more times at the same budget, while on long-horizon games it beats a search baseline \citep{wang2026rethinkingharness};} and that writing a harness update differs from benefiting from it \citep{harness_updating_not_benefit}. Each measures success; \ours{} also measures what a skill changes in the agent.}

\paragraph{Reruns and selection.} {Repeated selection on a fixed holdout overfits it \citep{dwork_reusable_holdout_2015,cawley2010overfitting}, and the best of several noisy estimates is biased upward, the optimizer's curse, the winner's curse of Section~\ref{sec:intro} \citep{smith_winkler_2006}. Agent evaluations thus call for repeated trials, error bars and a statement of the harness used \citep{yao2024taubench,miller2024errorbars,kapoor2024agentsmatter,hal,zhang2026stopcomparing}.}

\paragraph{Cost accounting.} {Agents of similar accuracy differ widely in cost \citep{kapoor2024agentsmatter,hal}. In agentic coding, input tokens drive the cost and vary from run to run \citep{token_spend}, and cutting tokens need not cut cost, because it changes the trajectory \citep{token_not_cost}. Attribution methods trace cost or failures to steps \citep{cost_utility,causal_agent_replay}. We trace the cost of a skill to the turns its rules add, each of which re-sends the history.}

\section{Problem Formulation}
\label{sec:setting}

\begin{figure*}[t]
\centering
\includegraphics[width=\textwidth]{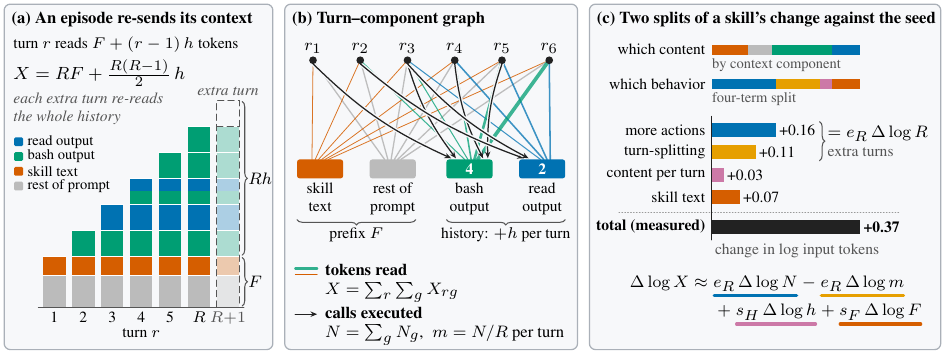}
\caption{\textbf{\ours{} splits a skill's extra input tokens by content and by behavior.} (a) Each turn re-sends the prefix and the history. (b) Turns linked to the context components they read or call. (c) The two splits against the seed.}
\label{fig:method}
\end{figure*}

{A \emph{skill} $S$ is text that the harness places in the model's context at the start of every episode. An \emph{episode} $e=(S,i,j)$ is attempt $j$ of skill $S$ on task $i$; it ends with an outcome $y\in\{0,1\}$ after $R$ turns (one model request each) and $N$ tool calls executed by the harness; its input tokens $X$ sum what every turn sends to the model, as the provider reports them, cached tokens included.}

{A skill optimizer starts from a seed skill $S_0$ and proposes \emph{candidate} skills: an editor model reads the agent's episodes on training tasks and edits a \emph{parent}, the seed or an earlier candidate. The change between them is an \emph{edit}; a candidate identical to its parent is \emph{unchanged}. The gate scores a candidate against its parent on a dev set $V$ and \emph{keeps} the edit when the optimizer's acceptance rule accepts it; a kept skill's \emph{margin} $\delta_t$ is its dev score minus its parent's. The optimizer returns the \emph{learned skill} $S$ it selected by success on $V$, its \emph{returned skill}, and reports the \emph{dev gain} $G=\sum_t\delta_t$ from the seed to the returned skill.}

{A \emph{test run} is one pass of a skill over all tasks of a test set $\mathcal I$. We run each skill $k$ times, so every task has $k$ episodes per skill, $j=1,\dots,k$. Success is the mean outcome, $Q(S)=\frac{1}{k|\mathcal I|}\sum_{i,j}y_{ij}(S)$ (avg@$k$). Cost is paired task by task against the seed:
\begin{equation}\label{eq:cost}
L(S)=\frac{1}{|\mathcal I|}\sum_{i\in\mathcal I}\log\frac{\bar X_i(S)}{\bar X_i(S_0)},\qquad\rho=\exp L(S),
\end{equation}
where $\bar X_i$ is the mean of $X$ over a skill's $k$ episodes of task $i$, and $\mathcal I$ keeps the tasks with outcomes and tokens recorded for both skills. Pairing cancels task length, and the log weighs long and short tasks alike. The same ratio over turns and tool calls gives $\rho^{\rm turns}$ and $\rho^{\rm calls}$. The \emph{input-token ratio} $\rho$ is the typical per-task ratio to the seed's; $\rho>1$ means the skill reads more.}

{We ask two questions of a learned skill against its seed: (i) does it succeed more often, that is, does its test gain exceed the spread of the seed's own test runs, and does its dev gain $G$ exceed what selection alone reports; and (ii) which content and which behavior make up its extra input tokens, $L(S)$.}

\section{Methodology}
\label{sec:attribution}

{\ours{} answers both questions from the episode logs, without changing the optimizer. For (i), it replays the optimizer's acceptance rule on unchanged candidates (Section~\ref{sec:replay}). For (ii), it records each episode as a bipartite graph of turns and the context they read, which splits $L(S)$ by content (Section~\ref{sec:graph}), and, through the turn elasticity (Section~\ref{sec:elasticity}), by behavior (Section~\ref{sec:split}; Figure~\ref{fig:method}).}

\subsection{Selection Replay}
\label{sec:replay}

{How the margins add up depends on the gate. A gate that scores every kept skill on one fixed dev set, as SkillOpt and GEPA do, lets the scores in between cancel, so $G$ is the returned skill's dev score minus the seed's \emph{starting score}, the dev score the run starts from. A gate that scores each step on a small batch, where a task is worth more points, and rescores the kept skill at the next step, as SkillAdam does, does not let the scores cancel, so the sum keeps every lucky win.

Because the gate keeps the best of several noisy dev scores and measures it against one noisy starting score, even unchanged candidates show a gain; the replay measures how large. Each dev task $v\in V$ has a difficulty $p_v$, the probability that a seed episode solves it, drawn from a prior $\mathrm{Beta}(a,b)$ fitted by maximum likelihood to the seed's dev and test episodes. If the seed solved $v$ in $s_v$ of its $n_v$ dev episodes, the difficulty of $v$ has the posterior
\begin{equation}\label{eq:replay}
p_v\mid\text{seed}\sim\mathrm{Beta}(a+s_v,\;b+n_v-s_v).
\end{equation}
A replay draws every $p_v$ from Eq.~\eqref{eq:replay}, scores every candidate as the seed scored again, and applies the optimizer's acceptance rule; the gain it reports is $G_0$. $P(G_0\ge G)$ is the share of replays reporting at least $G$; a large $P$ means $G$ is no evidence of improvement. The replay assumes episodes independent given $p_v$ (Appendix~\ref{app:assumptions}); Appendix~\ref{app:selection} gives the acceptance rules.}

\subsection{Episodes as Bipartite Graphs}
\label{sec:graph}

{We record each episode as a bipartite graph between its turns $r=1,\dots,R$ and its \emph{context components} $\mathcal C$, the labeled parts of what the model reads in a turn: the skill text, the rest of the prompt (system prompt, tool schemas and task), the agent and user messages, and each tool's outputs. A \emph{context edge} $(r,g)$ carries the $X_{rg}$ tokens of component $g$ that turn $r$ reads; a \emph{call edge} $(r,g)$ carries the $N_{rg}$ calls of tool $g$ that turn $r$ issues, whose outputs form component $g$ (Figures~\ref{fig:teaser} and~\ref{fig:method}b). The model reads $X=\sum_{r,g}X_{rg}$ tokens and the harness executes $N=\sum_{r,g}N_{rg}$ calls.}

{We split each turn's reported total over the components by content length so that $\sum_g X_{rg}=X_r$ holds exactly (Appendix~\ref{app:components}); the component totals $X_g=\sum_r X_{rg}$, which sum to $X$, say \emph{which content} the extra tokens are. Because the harness re-sends earlier content at every turn (Section~\ref{sec:elasticity}), three per-turn quantities describe how much is read: the \emph{prefix} $F$, the mean tokens per turn of the skill and the rest of the prompt; the \emph{content per turn} $h$, the tokens the conversation gains per turn; and the \emph{calls per turn} $m=N/R$.}

\subsection{Turn Elasticity}
\label{sec:elasticity}

{A harness that does not compact the history by summarizing old turns re-sends the prefix and the whole conversation at every turn. If the conversation grows by $h$ tokens per turn, turn $r$ reads $F+(r-1)h$ tokens (Figure~\ref{fig:method}a), and an episode of $R$ turns reads
\begin{equation}\label{eq:mechanism}
X=RF+\tfrac{R(R-1)}{2}\,h .
\end{equation}
We read $R$, $X$ and $F$ from the graph and define $h$ by Eq.~\eqref{eq:mechanism}: the content per turn of an episode whose history grew evenly. With the prefix share $s_F=RF/X$ and the history share $s_H=1-s_F$, a change of skill on the same task moves the input tokens by
\begin{equation}\label{eq:elasticity}
\Delta\log X\approx e_R\,\Delta\log R+s_H\,\Delta\log h+s_F\,\Delta\log F,
\end{equation}
where $\Delta$ is the difference between the two skills, the shares and $R$ are taken at the mean of the two skills' values, and
\begin{equation}\label{eq:eR}
e_R=1+s_H\,\frac{R}{R-1}\qquad(R\ge2)
\end{equation}
is the \emph{turn elasticity}: the percent change in input tokens per percent more turns, at fixed content per turn and prefix.}

\begin{proposition}[Cost of an extra turn]\label{prop:elasticity}
Let $R\ge2$, $F>0$ and $h>0$. Then $e_R>1$; $e_R$ increases with the history share $s_H$; $e_R<2$ if and only if a turn adds less than twice the prefix, $h<2F$; and $e_R\to2$ as $R$ grows at fixed $F$ and $h$. Without history, $h=0$ and $e_R=1$.
\end{proposition}

{An extra turn re-sends the prefix, so 1\% more turns adds at least 1\% more input tokens, and re-reads the history, so in a long episode it adds nearly 2\% (proof in Appendix~\ref{app:proofs}). Across episodes of one task, the regression of $\log X$ on $\log R$, the \emph{turn slope}, estimates the same elasticity with content per turn free to move with the turns.}

\subsection{The Four-Term Split}
\label{sec:split}

{Turns are calls over calls per turn, $\Delta\log R=\Delta\log N-\Delta\log m$, so Eq.~\eqref{eq:elasticity} splits a skill's change in input tokens against the seed into four terms, one per behavior:
\begin{equation}
\begin{split}
\Delta\log X\approx{}&\underbrace{e_R\,\Delta\log N}_{\text{more actions}}\;\underbrace{-\,e_R\,\Delta\log m}_{\text{turn-splitting}}\\
&+\underbrace{s_H\,\Delta\log h}_{\text{content per turn}}+\underbrace{s_F\,\Delta\log F}_{\text{skill text}}.
\end{split}
\label{eq:split}
\end{equation}
\emph{More actions} means more tool calls; \emph{turn-splitting} spreads the same calls over more turns, which issuing independent calls in one turn undoes; \emph{content per turn} counts more text entering the conversation each turn, such as longer tool outputs; \emph{skill text} is the longer prefix. The first term changes what the agent does, the other three how much it reads to do it.}

{The first two terms, the turn terms, sum to $e_R\,\Delta\log R$, the cost of extra turns. Content per turn is counted per turn, so the same history spread over more turns lowers it and offsets part of the turn-splitting term (Figure~\ref{fig:teaser}c). We evaluate Eq.~\eqref{eq:split} per task on the means over each side's $k$ episodes and average over tasks; the terms recover $L(S)$ of Eq.~\eqref{eq:cost} up to the first-order remainder of Eq.~\eqref{eq:elasticity}. Eq.~\eqref{eq:split} is an accounting split: it says which behavior the extra tokens come with.}

\section{Experiments}
\label{sec:findings}

{We ask three research questions about skills on the model and benchmark they were learned on. \textbf{RQ1}: Do learned skills succeed more often than the seed? \textbf{RQ2}: What do learned skills change in the agent's behavior, and what does that cost? \textbf{RQ3}: What in the optimizers' edits produces these changes? Section~\ref{sec:main} answers RQ1 and RQ2 and tests whether the skills transfer to another model or benchmark; Section~\ref{sec:analysis} answers RQ3 and traces RQ2's cost to its source.}

\subsection{Experimental Setup}
\label{sec:setup}

\begin{table*}[t]
\caption{\textbf{Test results} (avg@3). $Q$: success (\%), small: change from the seed in points; $\rho$, $\rho^{\rm turns}$, $\rho^{\rm calls}$: input tokens, turns and tool calls relative to the seed (Eq.~\eqref{eq:cost}), greener: lower. Bold, underline: best, second-best $Q$.}
\label{tab:main}
\centering
\footnotesize
\setlength{\tabcolsep}{3pt}
\newlength{\relqw}\settowidth{\relqw}{\scriptsize$-11.1$}
\newcommand{\relq}[1]{\,\makebox[\relqw][r]{\scriptsize$#1$}}
\newcommand{\relqnone}{\,\makebox[\relqw]{}}
\resizebox{\textwidth}{!}{%
\begin{tabular}{@{}l*{12}{r}@{}}
\toprule
\multirow{2}{*}{Skill} & \multicolumn{4}{c}{TB2} & \multicolumn{4}{c}{\taubench{}} & \multicolumn{4}{c}{AutoBench} \\
\cmidrule(lr){2-5} \cmidrule(lr){6-9} \cmidrule(lr){10-13}
 & \multicolumn{1}{c}{$Q$} & \multicolumn{1}{c}{$\rho$} & \multicolumn{1}{c}{$\rho^{\rm turns}$} & \multicolumn{1}{c}{$\rho^{\rm calls}$} & \multicolumn{1}{c}{$Q$} & \multicolumn{1}{c}{$\rho$} & \multicolumn{1}{c}{$\rho^{\rm turns}$} & \multicolumn{1}{c}{$\rho^{\rm calls}$} & \multicolumn{1}{c}{$Q$} & \multicolumn{1}{c}{$\rho$} & \multicolumn{1}{c}{$\rho^{\rm turns}$} & \multicolumn{1}{c}{$\rho^{\rm calls}$} \\
\midrule
\multicolumn{13}{c}{\textit{DeepSeek-V4.1-Flash}} \\
No skill & 72.2\relq{-3.7} & \heat{0.94} & \heat{1.00} & \heat{1.00} & 73.8\relq{-2.7} & \heat{0.98} & \heat{1.03} & \heat{1.01} & 36.4\relq{-8.6} & \heat{0.88} & \heat{0.92} & \heat{0.94} \\
\midrule
GEPA & \underline{74.1}\relq{-1.9} & \heat{1.34} & \heat{1.18} & \heat{1.21} & 75.6\relq{-0.9} & \heat{1.15} & \heat{1.07} & \heat{1.06} & 44.2\relq{-0.8} & \heat{1.21} & \heat{1.09} & \heat{1.00} \\
SkillOpt & 64.8\relq{-11.1} & \heat{1.43} & \heat{1.22} & \heat{1.23} & \textbf{79.1}\relq{+2.7} & \heat{1.22} & \heat{1.04} & \heat{1.09} & 41.7\relq{-3.3} & \heat{1.25} & \heat{1.11} & \heat{1.04} \\
SkillAdam & 68.5\relq{-7.4} & \heat{1.37} & \heat{1.22} & \heat{1.27} & 75.1\relq{-1.3} & \heat{1.09} & \heat{1.03} & \heat{1.07} & \underline{44.7}\relq{-0.3} & \heat{1.08} & \heat{1.03} & \heat{0.97} \\
\midrule
Seed & \textbf{75.9}\relqnone & \heat{1.00} & \heat{1.00} & \heat{1.00} & \underline{76.4}\relqnone & \heat{1.00} & \heat{1.00} & \heat{1.00} & \textbf{45.0}\relqnone & \heat{1.00} & \heat{1.00} & \heat{1.00} \\
\midrule
\multicolumn{13}{c}{\textit{GPT-6-Luna}} \\
No skill & \underline{70.6}\relq{+2.0} & \heat{0.96} & \heat{1.00} & \heat{0.98} & 64.9\relq{-4.4} & \heat{0.98} & \heat{1.05} & \heat{1.05} & 27.5\relq{-9.7} & \heat{0.82} & \heat{0.90} & \heat{0.92} \\
\midrule
GEPA & 68.6\relq{0.0} & \heat{1.00} & \heat{1.00} & \heat{1.00} & \underline{65.3}\relq{-4.0} & \heat{1.04} & \heat{1.02} & \heat{1.00} & 41.4\relq{+4.2} & \heat{1.07} & \heat{1.04} & \heat{1.11} \\
SkillOpt & 68.6\relq{0.0} & \heat{1.00} & \heat{1.00} & \heat{1.00} & 63.1\relq{-6.2} & \heat{1.01} & \heat{1.00} & \heat{0.99} & \underline{42.8}\relq{+5.6} & \heat{1.25} & \heat{1.18} & \heat{1.08} \\
SkillAdam & \textbf{74.5}\relq{+5.9} & \heat{1.20} & \heat{1.23} & \heat{1.20} & 64.9\relq{-4.4} & \heat{1.05} & \heat{1.06} & \heat{1.05} & \textbf{43.6}\relq{+6.4} & \heat{1.32} & \heat{1.24} & \heat{1.05} \\
\midrule
Seed & 68.6\relqnone & \heat{1.00} & \heat{1.00} & \heat{1.00} & \textbf{69.3}\relqnone & \heat{1.00} & \heat{1.00} & \heat{1.00} & 37.2\relqnone & \heat{1.00} & \heat{1.00} & \heat{1.00} \\
\bottomrule
\end{tabular}}
\end{table*}

\paragraph{Skill optimizers.} {We run three published general skill optimizers (Appendix~\ref{app:setup}). GEPA \citep{gepa} has its editor rewrite a parent after reading the trajectories of a batch of 8 training tasks (its default is 3) and scores a candidate on the whole dev set if it beats its parent on that batch. SkillOpt \citep{skillopt} keeps a candidate whose score on the whole dev set is strictly above the current skill's. SkillAdam \citep{skilladam} proposes small edits and keeps one only if it beats the current skill on 10 dev tasks drawn anew at each step; TB2 reuses the same 9. Under GPT-6-Luna, SkillAdam's gate uses the 10 training tasks its editor read in place of these dev tasks. They share one seed, \emph{verification-before-completion} \citep{superpowers} (Appendix~\ref{app:seed}); one sample of 160 training episodes per benchmark, from which they draw their training batches; and one dev set of 30 tasks (9 on TB2).}

\paragraph{Benchmarks, harness and models.} {Terminal-Bench 2.1 \citep{merrill2026terminalbench} (TB2) checks terminal tasks with test scripts. \taubench{} \citep{yao2024taubench,barres2025tau2bench,shi2026tauknowledge} pairs customer-service tasks with a simulated user. AutomationBench \citep{shepard2026automationbench} (AutoBench) poses business workflows, solved only when every programmatic check of the final state passes. The test sets hold 18, 75 and 120 tasks, disjoint from training and dev. The agent is DeepSeek-V4.1-Flash \citep{deepseek2026v41flash} or GPT-6-Luna (DeepSeek and Luna below) in OpenCode \citep{opencode}, a generalist harness with file and shell tools, run without history compaction. The three optimizers learn skills for each model; under Luna on TB2, GEPA and SkillOpt return the seed, which leaves sixteen learned skills.}

\paragraph{Metrics.} {Every skill, including the seed and the no-skill agent, runs three times on the test set, $k=3$. $Q$ is the mean success over these runs (avg@3); $\rho$, $\rho^{\rm turns}$ and $\rho^{\rm calls}$ follow Eq.~\eqref{eq:cost}; the four-term split is computed only on the tasks where all six episodes of both skills are \emph{usable}, with at least two turns, one tool call and positive content per turn.}

\subsection{Main Results}
\label{sec:main}\label{sec:tax}\label{sec:replicate}

\begin{finding}[RQ1]Learned skills do not succeed more often than the seed.\end{finding}

Averaged over every test task of all sixteen learned skills (Appendix~\ref{app:assumptions}), test success is $0.1$ points above the seed's, and $1.4$ below over the nine DeepSeek skills, which range from $-11.1$ to $+2.7$ points (Table~\ref{tab:main}). Under Luna the seven skills average $+1.6$ points over the seed: 4.0 to 6.2 points below it on \taubench{}, 4.2 to 6.4 above on AutoBench, and on TB2 GEPA and SkillOpt return the seed and SkillAdam scores 5.9 above it. From a three-word seed instead, Luna's SkillOpt and GEPA raise test success on AutoBench by 18.3 and 19.4 points, but only to the level of their runs from the hand-written seed, and gain nothing on \taubench{} (Appendix~\ref{app:scratch}).

\begin{finding}[RQ1]Selection alone can produce the dev gains the optimizers report.\end{finding}

\begin{figure*}[t]
\centering
\includegraphics[width=\textwidth]{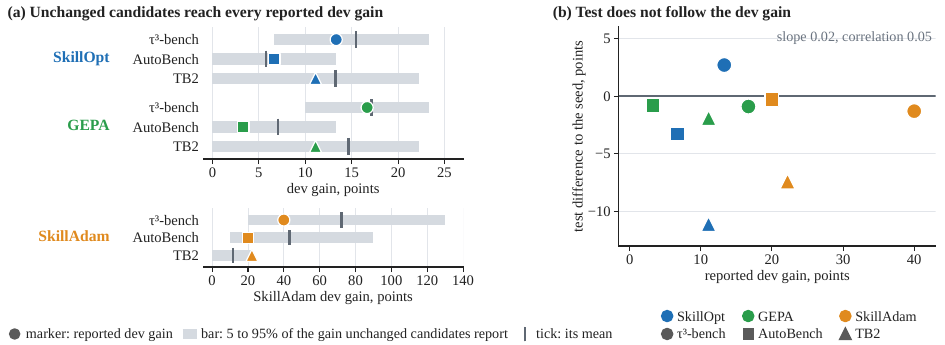}
\caption{\textbf{Reported dev gains are what selection alone reports, and test does not follow them.} DeepSeek-V4.1-Flash. (a) Each run's reported dev gain against replays with all candidates unchanged. (b) Test difference to the seed against the reported dev gain, nine returned skills (Table~\ref{tab:main}). SkillAdam's dev gain sums its accepted edits' margins on small batches, so it has its own axis in~(a). Replays: Appendix~\ref{app:selection}.}
\label{fig:devtest}\label{fig:devdyn}\label{fig:devchance}
\end{figure*}

Under DeepSeek, SkillOpt and GEPA reported dev gains of 3 to 17 points, and SkillAdam 20 to 40, summed over small dev batches (Section~\ref{sec:replay}). Unchanged candidates reach each of the nine reported gains with probability 0.25 to 0.94 (Figure~\ref{fig:devtest}a), and test does not follow the gains (Figure~\ref{fig:devtest}b): along the path of skills each \taubench{} run kept in turn, tested once on an updated DeepSeek release, the dev gain climbs 6.7 points after the first kept skill while the test difference stays within 2.7 points of its value there (Figure~\ref{fig:pathtest}, Appendix~\ref{app:version}). Every DeepSeek SkillOpt and GEPA run measures its gain against the seed's lowest-scoring dev run, one of two or three, its starting score, and keeps the best of 4 to 8 candidates on a dev set of at most 30 tasks. A gate that accepts any candidate that outscores its parent in one dev run accepts an unchanged candidate 24 to 41\% of the time, and replaying each optimizer's acceptance rule when every candidate is unchanged, that is, the seed scored again, reaches the reported gains with probability 0.56 to 0.94 for SkillOpt and GEPA under DeepSeek and 0.21 to 0.99 under Luna. Task by task, the ten SkillOpt and GEPA skills newly solve 12.5\% of the dev tasks and newly fail 2.7\%, but on test 8.9\% and 6.9\%, close to the 1.9 to 7.8\% of tasks on which two test runs of the seed itself differ in each direction, except Luna on AutoBench, where the learned skills gain on test although their dev scores hardly moved (Appendix~\ref{app:selection}).

\begin{finding}[RQ2]Learned skills sharpen the agent's behavior; they do not expand it.\end{finding}

An agent \emph{covers} a task if one of its three test runs solves it; a skill \emph{expands} the agent if it covers tasks the agent did not, and \emph{sharpens} it if it solves more often, with more similar tool sequences, tasks the agent already solved sometimes. Over the six model--benchmark pairs with learned skills, the seed covers 36 tasks that the no-skill agent does not and misses 13 that it covers, while the learned skills, summed over the three optimizers, cover 57 tasks that the seed does not and lose 62 that it covers. Where they gain, a task's three episodes agree more: tool sequences more alike than under the seed for 68\% of the tasks whose success rose, against 49\% of the unchanged (Appendix~\ref{app:dynamics}).

\begin{finding}[RQ2]Every skill learned under DeepSeek costs more than the seed.\end{finding}

All nine read 1.08 to 1.43 times the seed's input tokens per task and take 1.03 to 1.22 times its turns (Table~\ref{tab:main}). For eight of the nine, every test run reads more input tokens than any run of the seed, and the increase holds on the tasks that both skills solve in every test run (1.08 to 1.29), so it does not come from solving different tasks (Appendix~\ref{app:paired}). Averaged over the nine skills' test tasks, the input-token ratio $\rho$ is 1.18; over the tasks of the seven learned under Luna it is 1.14, and over all sixteen 1.16.

\begin{finding}Learned skills do not transfer across model or benchmark; most of their cost does.\end{finding}

We run Luna on the three \taubench{} skills learned under DeepSeek, and each DeepSeek TB2 or AutoBench skill once on the other benchmark. One of the nine is above the seed, by 1.9 points: under Luna the three score 3.6 to 6.2 points below it, as the skills learned under Luna do, and on the other benchmark five of the six score 2.5 to 9.3 points below it. The cost moves with the skill: under Luna the skills learned under DeepSeek read 1.08 to 1.38 times the seed's input tokens and those learned under Luna 1.01 to 1.05, and across benchmarks TB2's skills read 1.24 to 1.49 times the seed's on AutoBench and AutoBench's 1.00 to 1.22 on TB2 (Appendix~\ref{app:ood}).

\subsection{Additional Analysis}
\label{sec:analysis}\label{sec:where}\label{sec:text}\label{sec:robust}

One chain, from the editor through the agent to the harness, produces the extra cost. The gate scores a whole edit on at most 30 tasks and cannot credit a single rule, so the rules the editor writes accumulate.

\begin{finding}[RQ3]The editor adds rules, and none asks the agent to read less.\end{finding}

A \emph{rule} is one added bullet, numbered item, table row or example line; we sort each into one of eight categories by what it asks the agent to do, such as \emph{verify}, re-check a result, or \emph{serialize}, act one call at a time. Under DeepSeek the editor writes from small batches of training trajectories, an optimizer reads at most 160, yet a skill gains 37 rules on average, up to 105. Of the 59 DeepSeek edits scored on dev tasks whose parent the logs identify (38 under Luna), 58 lengthen the skill (33 under Luna); of the 337 rules they add (81 under Luna), none asks the agent to read less, half ask it to verify and 18 serialize (Appendix~\ref{app:growth}). Nor does the gate filter cost: kept and rejected edits raise input tokens over their parents by about 10 and 14\% on average.

\begin{finding}[RQ3]The agent carries the rules out as extra turns.\end{finding}

\begin{figure}[t]
\centering
\includegraphics[width=\linewidth]{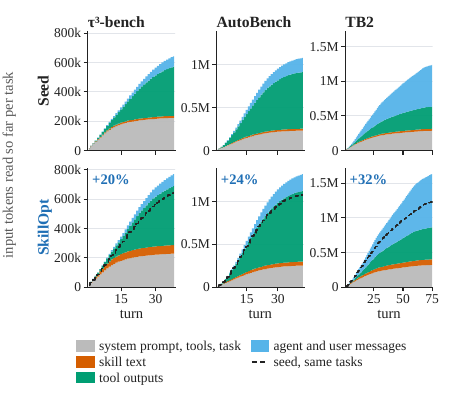}
\caption{\textbf{SkillOpt reads 20 to 32\% more input per task than the seed (DeepSeek-V4.1-Flash).} Stacked by context component, mean over tasks. Dashed: the seed's total on the same tasks; label: SkillOpt's total over the seed's, a ratio of totals, not $\rho$.}
\label{fig:cummain}
\end{figure}

The rules change the behavior they name. On AutoBench under DeepSeek, 56\% of the seed's writes, the calls that change state, share a turn with another write, against at most 0.3\% under each learned skill, whose writes go one per turn while reads stay grouped; these serialized writes are most of the turn-splitting there. Over the nine DeepSeek skills, extra turns carry 73\% of the summed log-token increase and the skill text 21\%; under Luna, too, extra turns outweigh the text. Figure~\ref{fig:cummain} shows SkillOpt under DeepSeek: the input read so far lies above the seed's, by 20, 24 and 32\% at the end, and the excess is mostly tool outputs, which the extra turns re-read, plus skill text on \taubench{}; on TB2 agent and user messages lead (Appendix~\ref{app:mech-turns}).

\begin{figure}[t]
\centering
\includegraphics[width=0.86\linewidth]{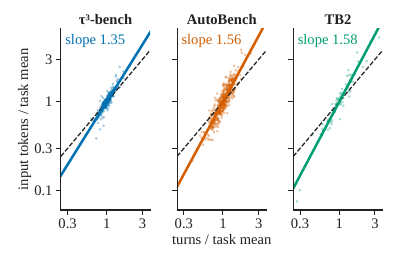}
\caption{\textbf{Input tokens grow faster than turns (DeepSeek-V4.1-Flash).} Dots: test episodes, one per skill and task; solid: within-task fit, dashed: slope 1.}
\label{fig:slopesin}
\end{figure}

\begin{finding}[RQ2]Input tokens grow faster than turns, because the harness re-sends the history.\end{finding}

As Proposition~\ref{prop:elasticity} predicts, within a task an episode with 10\% more turns reads about 14 to 16\% more input tokens (turn slopes 1.35 to 1.58, Luna 1.35 to 1.52; Figure~\ref{fig:slopesin}), while output tokens and tool calls rise at slopes of at most 1.03 under DeepSeek; the turn elasticity $e_R$ has medians 1.11, 1.76 and 1.64 on \taubench{}, AutoBench and TB2. Re-billing each recorded episode as if the harness kept only the last turn of history brings the slope to about 1 and leaves each skill at 1.05 to 1.36 times the seed's input tokens.

\section{Design Principles for Skill Optimizers}
\label{sec:implications}

The chain of Section~\ref{sec:analysis} shows where an optimizer can intervene: what its gate accepts, what it charges for, what its editor writes, and what the editor sees. One design principle follows for each (Figure~\ref{fig:implications}, Appendix~\ref{app:rq3}).

\begin{principle}Let the gate keep the parent unless the candidate gains more than selection alone reports. When success cannot separate the two, let cost decide, but only when the token difference exceeds what two runs of one skill differ by.\end{principle}

{A repeated win approximates this: in simulation, requiring a dev win to repeat on a second dev run accepts 6 to 17\% of unchanged candidates instead of 24 to 41\%, and keeps 55 to 94\% of edits that truly solve two more dev tasks (Figure~\ref{fig:implications}a). On test, every DeepSeek learned skill reads 1.08 to 1.43 times the seed's input tokens, while seven of the nine differ from the seed in success by less than the seed's runs do (Appendices~\ref{app:power} and~\ref{app:tiebreak}).}

\begin{principle}Charge each edit for the turns it adds, not for its length: count its extra turns against what it gains on dev.\end{principle}

{{Over the 59 DeepSeek edits whose parent the logs identify, the change in skill length explains 10\% of the variance in an edit's change in input tokens and the change in turns 93\% (Figure~\ref{fig:implications}b), and shorter skills can cost more \citep{xing2026skillremember}.} Loading the skill on demand instead of in every turn's prefix does not remove the cost: when the prompt says to load it, SkillOpt's \taubench{} skill reads 1.23 times the seed's input tokens either way (Appendix~\ref{app:delivery}).}

\begin{principle}Let the editor delete and merge rules.\end{principle}

{The three DeepSeek editors shorten the skill in 1 of 59 edits. Deleting the rules that split calls across turns would save less than it seems: recomputed with the learned skills at the seed's calls per turn, AutoBench's $\log\rho$ would halve, from 0.17 to 0.08 (Figure~\ref{fig:implications}c), but most of that comes from writes the rules serialize so that two writes in one turn cannot overwrite each other (Appendix~\ref{app:calls}). The editor also needs to write rules aimed at the costs that serialized writes do not explain: batch the independent reads that remain sequential, worth about one turn per episode, and return from each call only what the next turn needs.}

\begin{principle}Show the editor what its earlier rules cost: how many extra turns and input tokens they caused.\end{principle}

{Kept and rejected edits cost alike: 73 and 76\% of them raise input tokens over their parent (Figure~\ref{fig:implications}d). Only 5 of the 59 keep dev success and add no tokens, so a gate that required both would reject almost every edit; the cost has to be avoided as the edit is written.}

\section{Conclusion}
\label{sec:conclusion}

{Measured with \ours{}, the skills learned by GEPA, SkillOpt and SkillAdam sharpen the agent's behavior; they do not expand it. Over all sixteen learned skills of both models, test success is 0.1 points above the seed's; the skills cover about as many new tasks as they lose, make a task's episodes agree more where they gain, and read 1.16 times the seed's input tokens, over more turns. Selection alone can produce the dev gains the optimizers report, and the skills do not carry over to another model or benchmark, while most of their cost does. One causal chain explains the extra cost: on small dev sets the gate cannot credit single rules, so the rules the editor writes accumulate; the agent carries their checks out as extra turns; and input tokens grow faster than turns, because the harness re-sends the history. Four design principles follow (Section~\ref{sec:implications}); and to expand what an agent solves, an optimizer would need to reward coverage, the tasks solved in at least one of several episodes.}

\FloatBarrier
\section*{Limitations}

{\ours{} is a measurement instrument. We apply it to three published skill optimizers from one shared seed, so that their edits can be compared line by line, with two models in one harness on three benchmarks of long-horizon tasks; it applies unchanged to other optimizers, seeds, models, harnesses and task families. Three limits concern the instrument and one the optimizers it measures.}

{\paragraph{Post-hoc measurement.} \ours{} measures what a learned skill changed once its episodes are recorded: it attributes a change in input tokens to the behaviors that carry it, and a change in success to what selection alone reports, under the assumptions stated in Appendix~\ref{app:assumptions}. It does not predict which rule will add turns, and it does not explain why the model chooses a behavior; the link from a rule to the behavior it causes rests here on the edits the optimizers made and on matching rules to the behavior they name, such as serialized writes. Controlled rewrites of single rules would test this link causally, and a model of it would let an optimizer estimate an edit's cost before running it.}

{\paragraph{Harnesses that compact their history.} The turn elasticity is derived for a harness that re-sends its history, as OpenCode does in our runs. For harnesses that compact or summarize their history, or bill cached history at a discount, the bipartite graph and the four-term split still apply, but the turn elasticity has to be measured from their episodes; our re-billing of the recorded episodes under a compacted history, with the agent's turns held fixed, is a first step.}

{\paragraph{Decomposing success.} \ours{} decomposes cost because input tokens and tool calls add up over turns and context components. Success does not: it is scored once, at the end of an episode, and splitting it over turns, components or rules needs a value function over the agent's states. Behind a black-box model API the model's state is not observable, so no such value function can be defined, let alone decomposed. This is the credit a gate would need to judge single rules, and the most important extension of \ours{}.}

{\paragraph{Static optimizers and recursive self-improvement.} What separates these optimizers from recursive self-improvement (RSI) is not which artifact they optimize, a skill, a harness or a model, but whether the optimizer itself improves. An external optimizer does not solve tasks; its job is to make the agent better. Improving it continually means making ``improve the agent, starting from a given agent'' its training task, rather than training on ordinary task scores. Today's skill, harness and model optimizers are static propose-verify-accept loops: a fixed procedure proposes a change, scores it on tasks and keeps or discards it, and optimizing several artifacts together is still such a loop. \ours{} measures these loops and extends to harness and model optimizers, but they are not RSI. RSI needs every link of the chain to keep improving: the agent, the optimizer that improves it, and whatever improves that optimizer.}

\bibliography{references}

\appendix
\onecolumn
\etocdepthtag.toc{appendix}
\raggedbottom

\begingroup
\etocsettagdepth{main}{none}
\etocsettagdepth{appendix}{subsection}
\etocsettocstyle{\section*{Contents of the Appendices}\etocmarkboth{Contents of the Appendices}}{}
\tableofcontents
\endgroup
\clearpage

\section{\texorpdfstring{Experimental Setup}{Experimental Setup}}\label{app:setup}

{\paragraph{Benchmarks and splits.}\label{app:d} We use Terminal-Bench 2.1 (TB2); tau2-bench v1.0.1 \citep{yao2024taubench,barres2025tau2bench} with the task fixes of \citet{cuadron2026saber}, plus the $\tau$-Banking domain of $\tau$-Knowledge \citep{shi2026tauknowledge} (\taubench{}), whose simulated user runs on the agent's model; and the public set of AutomationBench (AutoBench). Each benchmark is split once into training, dev and test tasks in the ratio 7:1:2. The test set has 18 tasks on TB2, 75 on \taubench{} and 120 on AutoBench. Success is binary: a task is solved when its test scripts pass (TB2), when its reward is 1 under the official evaluator (\taubench{}), or when every check of the final state passes (AutoBench; 3 to 64 checks per task, median 11.5).}

{\paragraph{Harness and models.}\label{app:harness} The agent runs in OpenCode v1.14.31, with its released system prompt and control loop, dedicated file tools and a shell tool, and keeps the whole history. Both models, DeepSeek-V4.1-Flash \citep{deepseek2026v41flash} and GPT-6-Luna, run at high reasoning effort with at most 32k output tokens per turn. The skill follows the task in the first user message, never in the system prompt; the no-skill agent receives the task alone. On TB2 one test task whose test script timed out is excluded for GPT-6-Luna, leaving 17 of 18. The path tests of Appendix~\ref{app:version} and the loading checks of Appendix~\ref{app:delivery} use an updated release of DeepSeek-V4.1-Flash and take their baseline from runs on that release.}

{\paragraph{Skill optimizers.}\label{app:g2}\label{app:instantiations} Our baselines are the three published optimizers, each running its released code (Appendix~\ref{app:baselineprompts} gives the text we fill into their prompts). Each keeps or rejects candidates by success alone and returns a skill chosen on dev or training-batch scores, never test (SkillOpt's upstream loop also scores three skills on test; these scores select nothing). All three take their training batches in a fixed order from one shared sample of 160 training episodes (on TB2, three reshuffled passes over the 62 training tasks); a skill already run on a task reuses its stored episode.
\emph{GEPA} runs 20 rounds, so that its batches of 8 cover the 160 training episodes once, and returns the best on dev.
\emph{SkillOpt} makes one pass where upstream makes four, 4 rounds of 40 training tasks (on TB2 under DeepSeek, 5 rounds of 31), and three passes under GPT-6-Luna on TB2, the only run in which its slow update and meta-skill act; the slow update draws its own training tasks and enters the skill without a dev check. It returns the best on dev.
\emph{SkillAdam} takes 16 steps of 10 training tasks (on TB2 under DeepSeek, 18 steps on batches of 9, two of them of 8) and returns the last kept one.}

{\paragraph{Counting tokens, calls and components.}\label{app:components} Input tokens add up every turn's count as the provider reports it with the model's reply, cached tokens included; tool calls are the calls OpenCode executes.}
Each turn's input is then split over the context components of Section~\ref{sec:graph} in proportion to their content length, at four characters per token, and rounded so that the parts add up exactly. We call an episode \emph{usable} when it has at least two turns, one tool call and positive content per turn, $h>0$; only usable episodes enter the attribution.

\section{\texorpdfstring{Proof}{Proof}}\label{app:proofs}

{\paragraph{Proposition~\ref{prop:elasticity}.} Write Eq.~\eqref{eq:mechanism} as $\log X=\log R+\log\bigl(F+\tfrac{R-1}{2}h\bigr)$. At fixed $F$ and $h$,
\[
\frac{\partial\log X}{\partial\log R}=1+\frac{Rh/2}{F+(R-1)h/2}=1+\frac{R^2h/2}{X}=1+s_H\frac{R}{R-1},
\]
using $s_H=R(R-1)h/(2X)$; the partial derivatives with respect to $\log h$ and $\log F$ are $s_H$ and $s_F$, which gives Eq.~\eqref{eq:elasticity} to first order; the dropped higher-order part is the remainder reported in Appendix~\ref{app:mech-turns}. Since $R\ge2$ and $h>0$ give $s_H>0$, $e_R>1$; without history ($h=0$), $s_H=0$ and $e_R=1$. Moreover, at fixed $R$, $e_R$ increases with $s_H$, as Eq.~\eqref{eq:eR} shows directly. As $R\to\infty$ at fixed $F$ and $h>0$, $s_H\to1$ and $R/(R-1)\to1$, so $e_R\to2$. Finally, $e_R<2$ if and only if $Rh/2<F+(R-1)h/2$, that is $h<2F$.}

\section{\texorpdfstring{Selection and Significance}{Selection and Significance}}\label{app:rq1}\label{app:measurement}

Of the sixteen learned skills, only GPT-6-Luna's SkillAdam on AutoBench beats the seed on test by more than the seed's own run-to-run spread, 6.4 points against 5.8, and the reported dev gains are what selection alone reports, so they do not carry over to test.

\subsection{\texorpdfstring{Task Model and Test Intervals}{Task Model and Test Intervals}}\label{app:assumptions}

\begin{figure*}[t]
\centering
\includegraphics[width=0.92\textwidth]{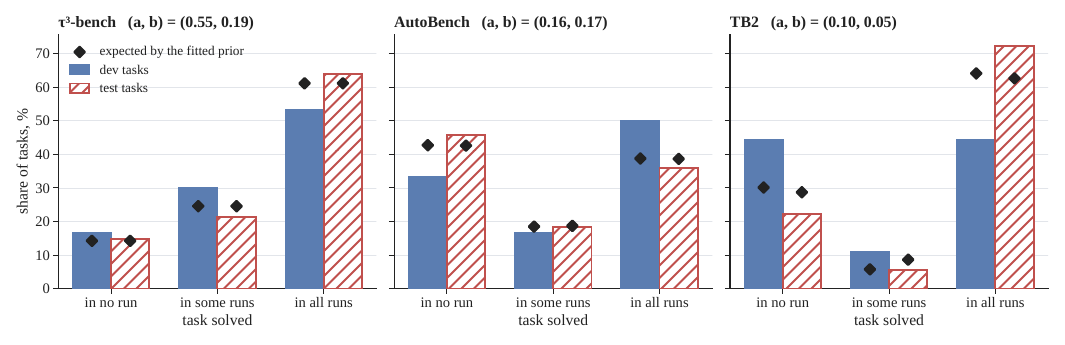}
\caption{\textbf{Most tasks are solved almost always or almost never, about as the fitted prior expects.} Bars: share of dev and test tasks the seed solves in none, some or all of its runs, DeepSeek-V4.1-Flash; diamonds: the share the fitted prior expects; titles: its $(a,b)$.}
\label{fig:priorfit}
\end{figure*}

Most tasks are solved almost always or almost never, so only a few can change a skill's score from run to run. Each task has a difficulty $p_v$, the probability that an episode of the seed solves it, and a benchmark's difficulties follow a prior $\mathrm{Beta}(a,b)$ fitted by maximum likelihood on the seed's dev and test episodes; no learned skill's episode enters the fit. Under DeepSeek-V4.1-Flash, $(a,b)=(0.10,0.05)$ on TB2, $(0.55,0.19)$ on \taubench{} and $(0.16,0.17)$ on AutoBench, all with $a,b<1$, a prior that puts most tasks near always or never solved. Counted over the seed's three test runs and its dev runs, two on TB2, three on \taubench{} and two or three on AutoBench, the tasks fall into about the shares this prior expects (Figure~\ref{fig:priorfit}). Only 1 of 18 test tasks on TB2, 16 of 75 on \taubench{} and 22 of 120 on AutoBench are solved in some but not all of their episodes. The figure checks the task model of the DeepSeek replay in Figure~\ref{fig:devtest}; under Luna the seed has one dev episode per task, so its dev tasks cannot be sorted this way.

The model assumes that once $p_v$ is fixed the episodes of a task are independent (A1), and that the difficulties of a benchmark's tasks are exchangeable draws from one Beta (A2), although domains differ in difficulty.

Pooled over skills, the learned skills do not succeed more often than the seed: over every test task of the nine DeepSeek skills the test difference is $-1.4$ points, with a one-sided 95\% upper bound of $+0.6$, and over all sixteen skills of both models $+0.1$ and $+1.6$. For one skill, the interval is a paired bootstrap over test tasks; for several, the per-task differences of all skills are averaged so that every test task of every skill counts once, and a cluster bootstrap resamples each benchmark's tasks, keeping a task's differences together. The input-token ratio $\rho$ over several skills is the same average taken on per-task log ratios, then exponentiated.

\subsection{\texorpdfstring{Run-to-Run Noise}{Run-to-Run Noise}}\label{app:power}

\begin{figure*}[t]
\centering
\includegraphics[width=\textwidth]{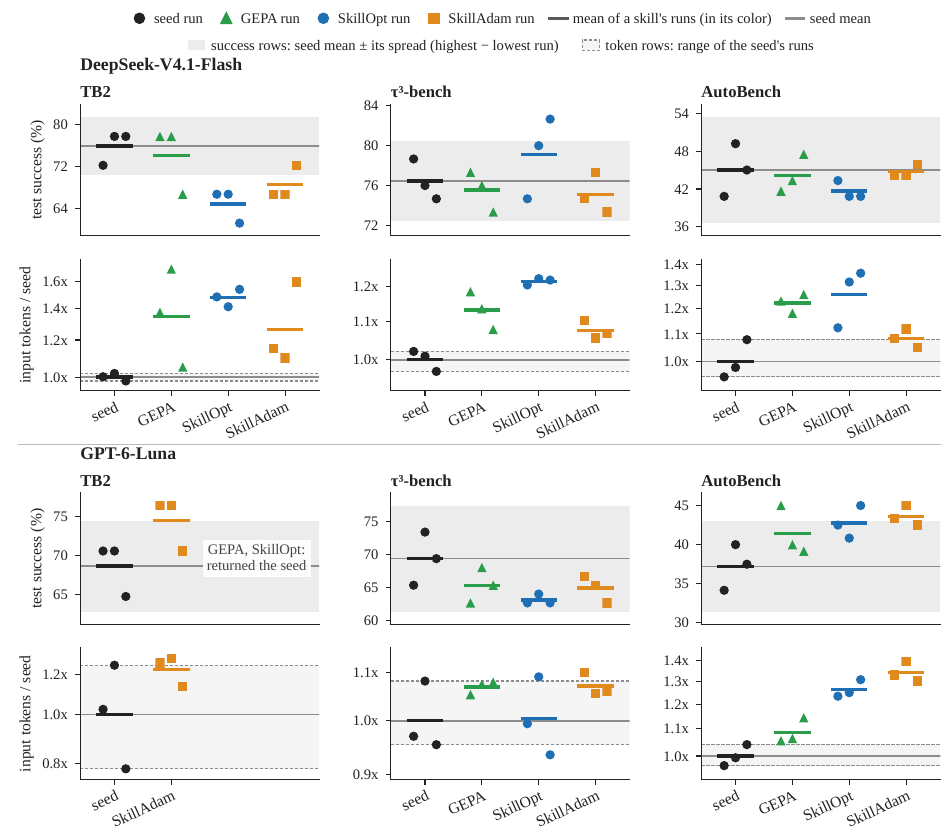}
\caption{\textbf{Most learned skills differ from the seed in success by less than the seed's own spread from run to run, but under DeepSeek-V4.1-Flash almost every one of them reads more input tokens than any run of the seed.} One point per test run, DeepSeek-V4.1-Flash above and GPT-6-Luna below; a mean of a skill's runs inside a success row's grey band differs from the seed by at most the seed's spread. A run's input tokens: per task, the ratio to the geometric mean of the seed's three runs, then the geometric mean over tasks; token-row means are geometric.}
\label{fig:reruns}
\end{figure*}

Most learned skills differ from the seed in success by less than the seed differs from itself, yet read more input tokens than any run of it (Figure~\ref{fig:reruns}). Under DeepSeek-V4.1-Flash the seed's spread, the gap between its highest and lowest test run, is 5.6 points on TB2, 4.0 on \taubench{} and 8.4 on AutoBench. Seven of the nine DeepSeek skills differ from the seed by less than the spread on their benchmark, all but SkillOpt and SkillAdam on TB2 ($-11.1$ and $-7.4$ points against 5.6), while for eight of the nine every run reads more input tokens than any run of the seed, all but SkillAdam on AutoBench (lowest run 1.05 against the seed's highest, 1.08). Under GPT-6-Luna the spreads are 8.0 points on \taubench{} and 5.8 on AutoBench, and the skills differ from the seed by $-4.0$ to $-6.2$ and $+4.2$ to $+6.4$ points; only SkillAdam's $+6.4$ on AutoBench exceeds the spread, and on TB2, where GEPA and SkillOpt returned the seed and so have no runs of their own, SkillAdam's $+5.9$ equals the seed's spread there, 5.9 points. Under GPT-6-Luna only the three AutoBench skills read more input tokens in every run than in any run of the seed.

\subsection{\texorpdfstring{Replaying Selection with Unchanged Candidates}{Replaying Selection with Unchanged Candidates}}\label{app:selection}

Unchanged candidates reach the reported dev gains. Under DeepSeek-V4.1-Flash, $P(G_0\ge G)$, estimated from 100k replays per optimizer run, is 0.56 to 0.94 across the SkillOpt and GEPA runs. Under GPT-6-Luna, where the seed has one dev episode per task, it is 0.47 to 0.99 on \taubench{} and 0.21 to 0.66 on AutoBench, over runs and four task models: the fitted one, a prior with fewer tasks near always or never solved, and two estimated from all dev episodes of the benchmark's optimizer runs. The reported dev gains are no evidence that any edit helped.

\phantomsection\label{app:checks}\label{app:variants}\label{app:devsize}The gains come from best-of-$K$ selection against a low starting score. Every DeepSeek SkillOpt and GEPA run started from the seed's lowest-scoring stored dev run; its other dev runs score 1 to 5 tasks above it, as much as the reported gains of $+1$ task each on TB2, $+4$ and $+5$ on \taubench{} and $+2$ and $+1$ on AutoBench. If the seed's baseline score is instead drawn at random from the task model, the mean replayed gain is only 0.6 to 2.3 tasks and the reported gains on \taubench{} become unlikely, with $P(G_0\ge G)$ between 0.02 and 0.17 across the fitted task model and two that estimate each task's rate from its own episodes.

Each replay draws every dev task's $p_v$ from its posterior, then one outcome per task for each of $K$ unchanged candidates, as many as the run scored (5, 4 and 4 for SkillOpt and 8, 7 and 7 for GEPA on TB2, \taubench{} and AutoBench), and applies the optimizer's acceptance rule. SkillOpt keeps the best candidate if it strictly beats the starting score; GEPA takes the best of the seed and the candidates and breaks ties for the seed; SkillAdam under DeepSeek compares at each step its current skill with unchanged candidates of that skill on 10 fresh dev tasks, on TB2 on the same 9 each step, and accepts the first that solves strictly more, drawing the current skill's outcome on a task once and keeping it.

Unchanged candidates also reach SkillAdam's dev gains, with probability 0.25, 0.87 and 0.88 on TB2, \taubench{} and AutoBench. Its kept edits won by 2 of 9 tasks on TB2, 22.2 points; by 3 and 1 of 10 on \taubench{}, which sum to 40.0 points; and by 2 of 10 on AutoBench, 20.0 points.

\begin{figure}[t]
\centering
\includegraphics[width=0.96\textwidth]{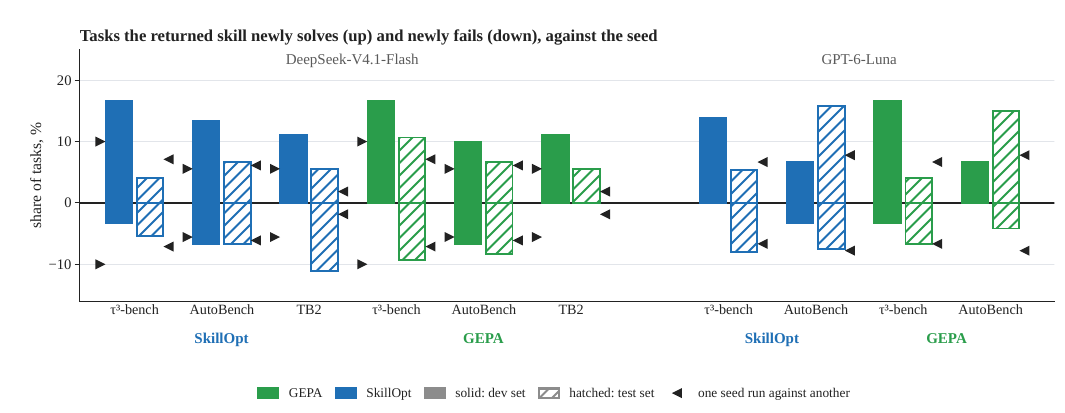}
\caption{\textbf{On dev the returned skill rarely loses a task the seed solved; on test it gains about as many tasks as it loses, except GPT-6-Luna on AutoBench, where it gains more.} Per optimizer run, the returned skill against the seed, both scored once inside that run on the whole dev set (solid bars) and on the whole test set (hatched bars); up: share of tasks the seed fails and the skill solves; down: the reverse; triangles: the share of tasks one run of the seed solves and another fails on the same set, averaged over pairs of its runs, so the same up and down; GPT-6-Luna's dev bars have none.}
\label{fig:devpertask}
\end{figure}

Task by task, the dev gains do not carry over to test (Figure~\ref{fig:devpertask}). Over the ten SkillOpt and GEPA skills, each scored once with the seed on the whole dev and test sets in its optimizer run, the returned skill newly solves 12.5\% of the dev tasks and newly fails 2.7\%, since the optimizer kept what scored higher there; on test it newly solves 8.9\% and newly fails 6.9\%. The figure compares only SkillOpt and GEPA, whose gates score every kept skill on one fixed dev set (Section~\ref{sec:replay}); SkillAdam's gate scores a small batch at each step, and under Luna on TB2 GEPA and SkillOpt return the seed. Without Luna on AutoBench the test shares are 6.2\% and 7.3\%, close to the share of tasks one test run of the seed solves and another of its runs fails, averaged over pairs of its three test runs (black triangles; under Luna the seed has one dev episode per task, so its dev bars have none). On Luna's AutoBench the test differences are large, 15.8\% and 15.0\% newly solved against 7.5\% and 4.2\% newly failed, while the dev set moved by one and two tasks.

\begin{figure}[t]
\centering
\includegraphics[width=\textwidth]{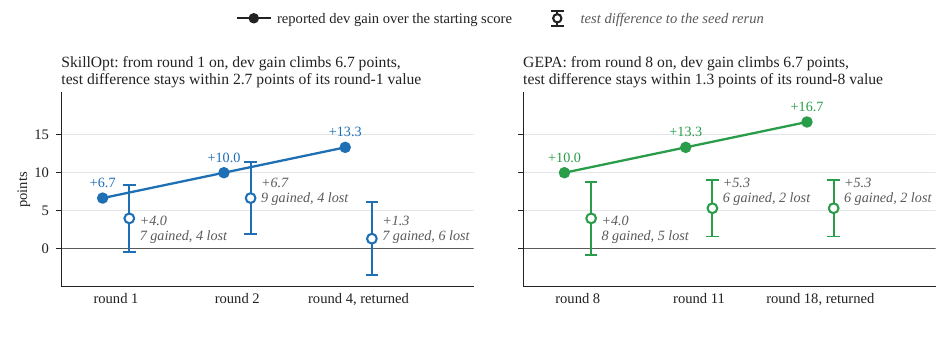}
\caption{\textbf{Along both \taubench{} paths, from the first kept skill on, the reported dev gain climbs 6.7 points while the test difference stays within 2.7 points of its value there.} DeepSeek; the test runs here, of the skills and of the seed rerun, are on an updated release, so only differences within this figure compare. Columns: the skills each run kept in turn as its best on dev; round: the optimizer iteration that wrote the skill; filled dots: reported dev gain over the starting score; open circles and italic grey labels: test difference to the seed rerun, whiskers $\pm1$ standard error paired by task; gained/lost: test tasks the skill solves and the seed rerun fails, and the reverse. The two series have different baselines, so compare how each changes along the path, not their levels.}
\label{fig:pathtest}
\end{figure}

\phantomsection\label{app:version}Along the \taubench{} paths, from the first kept skill on, the reported dev gain climbs 6.7 points while the test difference stays within 2.7 points of its value there for SkillOpt and within 1.3 for GEPA (Figure~\ref{fig:pathtest}). We test each of the six skills that became the best on dev in turn once on the 75 test tasks on the updated release of DeepSeek-V4.1-Flash, against a seed rerun on it. Of the DeepSeek runs, only those on \taubench{} kept a skill between the seed and the returned one: on TB2 and AutoBench the returned skill is the first to score above the seed on the dev set, and SkillAdam keeps or rejects each step on a small batch rather than on one dev set. The six lie $+1.3$ to $+6.7$ points above the seed rerun, but the rerun solves 5.8 points fewer tasks than the seed did in its three earlier runs, against which the six lie at $-4.4$ to $+0.9$ points, $-1.3$ on average.

\subsection{\texorpdfstring{Acceptance Rules}{Acceptance Rules}}\label{app:selrules}

\begin{figure}[H]
\centering
\includegraphics[width=\textwidth]{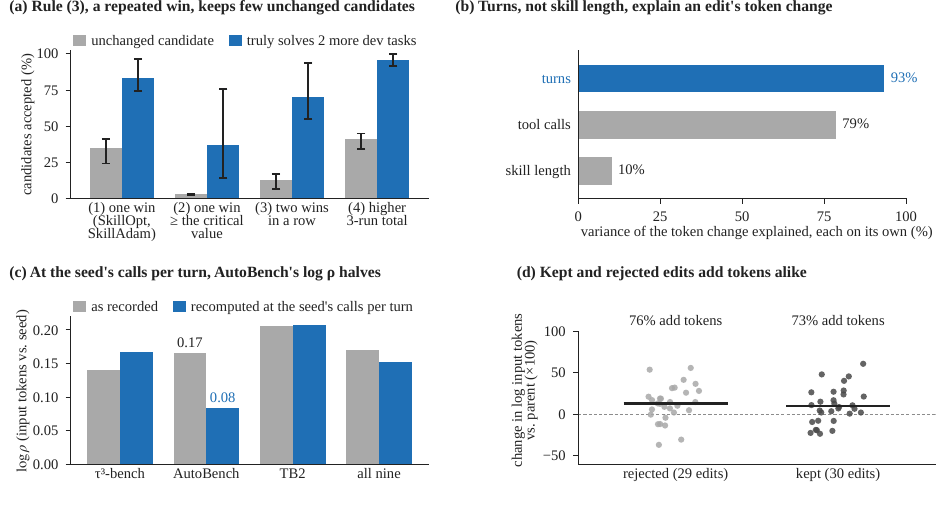}
\caption{\textbf{One panel per design principle of Section~\ref{sec:implications}, in order.} DeepSeek-V4.1-Flash. (a) Simulated acceptance rules (1) to (4) of Appendix~\ref{app:selrules}; grey: an unchanged candidate, blue: one that truly solves two more dev tasks; bar: mean over the three benchmarks, whiskers: lowest and highest benchmark. (b) Share of the variance of an edit's change in log input tokens that its change in turns, tool calls or skill length explains, each on its own. (c) Mean $\log\rho$ of a benchmark's three learned skills, or of all nine; grey: as recorded, blue: recomputed with each skill's calls packed into turns as densely as the seed's. (d) One dot per edit; short bar: mean, dashed line: zero.}
\label{fig:implications}
\end{figure}

Requiring a dev win to repeat refuses most unchanged candidates and keeps most real gains. We compare four rules on the task model, each deciding between one candidate and its parent (Figure~\ref{fig:implications}a, left to right). Rule (1) accepts a higher score in one dev run each, as SkillOpt and, under DeepSeek, SkillAdam do; it accepts 24, 41 and 39\% of unchanged candidates on TB2, \taubench{} and AutoBench. Rule (2) accepts a single-run win only if it reaches the critical value, the smallest margin an unchanged candidate reaches against its parent in at most 5\% of draws (2, 5 and 4 tasks); it accepts only 2.7 to 3.2\% of unchanged candidates but also only 14 to 21\% of true two-task gains on \taubench{} and AutoBench. Rule (3) requires the win to repeat on a second dev run: it accepts 6.4, 17.1 and 15.2\% of unchanged candidates and keeps 94, 55 and 62\% of true two-task gains. Rule (4), a higher total over three dev runs each, accepts 34 to 45\% of unchanged candidates, more than rule (1), because ties are rarer over three runs. Of the 59 DeepSeek edits whose parent the logs identify, the optimizers kept 30, and rule (2) would have refused 29 of the 30; these 29 raised log input tokens by 0.101 on average.

\label{app:tiebreak}Cost separates edits where success cannot. For an edit, the success $z$ is the summed per-task difference in success over the dev tasks both ran and the token $z$ the mean per-task log ratio of input tokens, each divided by its standard error; a change is \emph{beyond noise} when its $|z|$ is at least 1.96. The cost rule keeps the parent unless the candidate wins by rule (2)'s critical value or, without losing by it, reads fewer input tokens beyond noise. Of the 43 DeepSeek edits whose parent was scored in the same optimizer run, 9 change input tokens beyond noise and only 1 changes success; the optimizers kept 21 of the 43, of which 4 raised input tokens beyond noise and none won by rule (2)'s critical value. On test, too, the cost differences carry over and the success differences do not: every skill learned under DeepSeek reads 1.08 to 1.43 times the seed's input tokens, while success differs by $-11.1$ to $+2.7$ points (Table~\ref{tab:main}). Under GPT-6-Luna tokens separate only 2 of the 38 edits, so the cost rule mostly keeps the parent, returning the seed even instead of GEPA's and SkillOpt's AutoBench skills, which score above it on test.

\subsection{\texorpdfstring{Starting from a Three-Word Seed}{Starting from a Three-Word Seed}}\label{app:scratch}

\begin{figure*}[t]
\centering
\includegraphics[width=\textwidth]{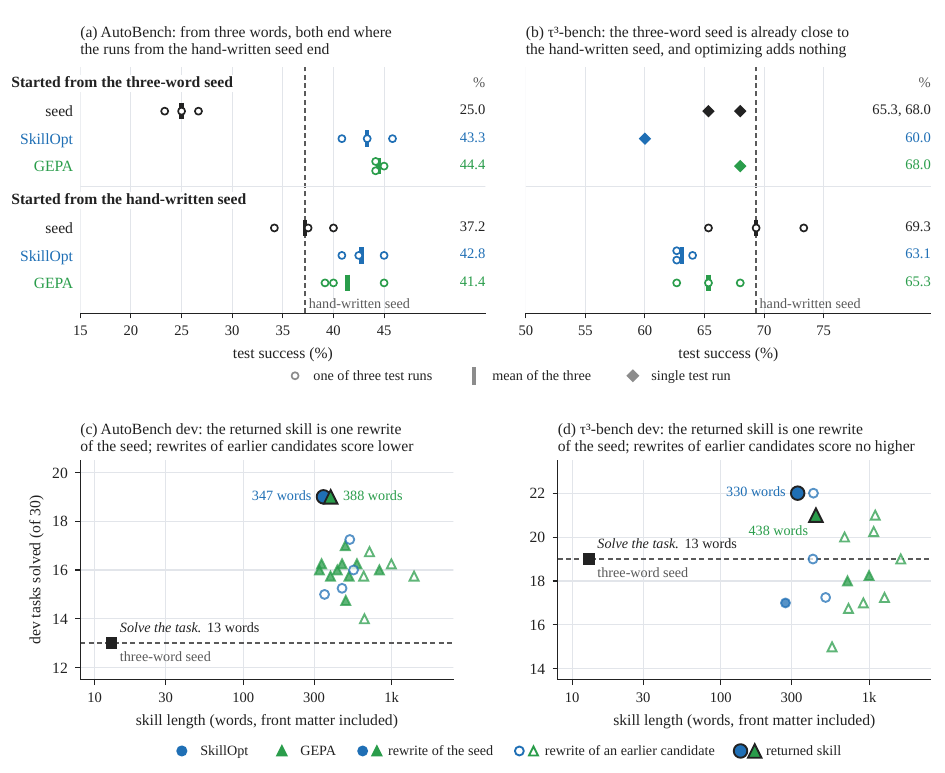}
\caption{\textbf{One rewrite of the three-word seed reaches the runs from the hand-written seed on AutoBench and adds nothing on \taubench{}; rewriting it again does not help.} (a,~b)~Test success of GPT-6-Luna; the number at right is the mean of the three test runs, or the single test runs. (c,~d)~Every candidate's dev score against its length; the dashed line is the three-word seed's dev score; marks that share a dev score are set slightly apart in height.}
\label{fig:scratch}
\end{figure*}

\textbf{Started from three words, the optimizers match the runs from the hand-written seed on AutoBench and add nothing on \taubench{}.} We run SkillOpt and GEPA under GPT-6-Luna from ``Solve the task.'' once each on AutoBench and \taubench{}, with the dev and test sets and the acceptance rules of the main runs (Figure~\ref{fig:scratch}). On AutoBench, test success rises from 25.0\% to 43.3\% under SkillOpt and 44.4\% under GEPA, 18.3 and 19.4 points, past the hand-written seed's 37.2\% and level with the runs started from it, 42.8\% and 41.4\%. On \taubench{}, where each optimizer run tests the three-word seed and its returned skill once, the three-word seed already scores 65.3\% and 68.0\% in its two test runs, close to the hand-written seed's 69.3\%; the returned skills score 60.0\% and 68.0\%, and their dev gains, $+10.0$ and $+6.7$ points, are what selection alone reports: replaying each optimizer's acceptance rule with unchanged candidates reaches them with probability 0.27 and 0.74.

\textbf{Each returned skill is one rewrite of the seed.} In both optimizers the editor rewrites a parent into a candidate, SkillOpt the skill it currently keeps and GEPA a skill that is best on at least one dev task, so every candidate rewrites either the seed or an earlier candidate. Both SkillOpt runs accept one candidate, and each GEPA run returns a rewrite of the seed (Figure~\ref{fig:scratch}c,d); every rewrite of a returned skill is longer and scores no higher on dev. The four returned skills have 347, 330, 388 and 438 words, against 13 for the three-word seed (every count includes the front matter, which holds the skill's name and description) and 580 for the hand-written seed; Appendix~\ref{app:scratch-skills} lists them.

\textbf{SkillOpt on AutoBench re-derives the seed's verification and adds serialized writes.} Three of its rules restate the hand-written seed: a successful response is not proof (the seed's ``verify independently''), read back before claiming completion (its Iron Law), and check a report against the requested fields (its line-by-line requirements check). One has no counterpart in the seed: ``perform mutations sequentially''. Behavior splits the same way. Under the hand-written seed, 91.1\% of AutoBench episodes follow their last write by a later read, against 53.8\% under the three-word seed and 97.2\% and 97.8\% under the learned skills; but 47.8\% of its writes, over all its 360 episodes, still share a turn with another write, against 58.7\% under the three-word seed and 4.3\% and 1.5\% under the learned skills. Serialized writes are the likely source of the 6.1 and 7.2 points by which the learned skills beat the hand-written seed on AutoBench; these single runs cannot separate that rule from the rules added with it.

\textbf{From three words, the learned skills mostly sharpen what Luna already solves without them.} On AutoBench the three-word seed scores like no skill, 25.0\% against 27.5\%. Of the 18.3 and 19.4 points the learned skills add, 16.7 each come from the 63 tasks that the no-skill agent or the hand-written seed solves in at least one run, and 1.7 and 2.8 from the other 57. Of the 35 tasks the three-word seed solves in one or two of its three runs, the learned skills solve 20 and 22 in all three.

\section{\texorpdfstring{Cost Analysis}{Cost Analysis}}\label{app:rq2}\label{app:mechanism}

\subsection{\texorpdfstring{Extra Cost on the Same Tasks}{Extra Cost on the Same Tasks}}\label{app:paired}\label{app:failcost}

Every skill learned under DeepSeek-V4.1-Flash reads more input tokens than the seed on the same test tasks: the paired bootstrap interval of $\rho$ lies above 1 for seven of the nine, and for eight the increase exceeds the spread of the seed's three test runs (Appendix~\ref{app:power}). The increase does not come from solving different tasks: on the tasks that a skill and the seed both solve in all three test runs, $\rho$ is 1.08 to 1.29 for all nine.

\subsection{\texorpdfstring{Where the Extra Tokens Come From}{Where the Extra Tokens Come From}}\label{app:mech-turns}\label{app:mech-attribution}

Input tokens grow faster than turns: the turn slope, fitted within tasks, is 1.35, 1.56 and 1.58 on \taubench{}, AutoBench and TB2 under DeepSeek (Figure~\ref{fig:slopesin}), against at most 1.03 for output tokens and tool calls, and the turn elasticity $e_R$ (Eq.~\eqref{eq:eR}) has medians 1.11, 1.76 and 1.64.

\begin{figure}[H]
\centering
\includegraphics[width=0.82\textwidth]{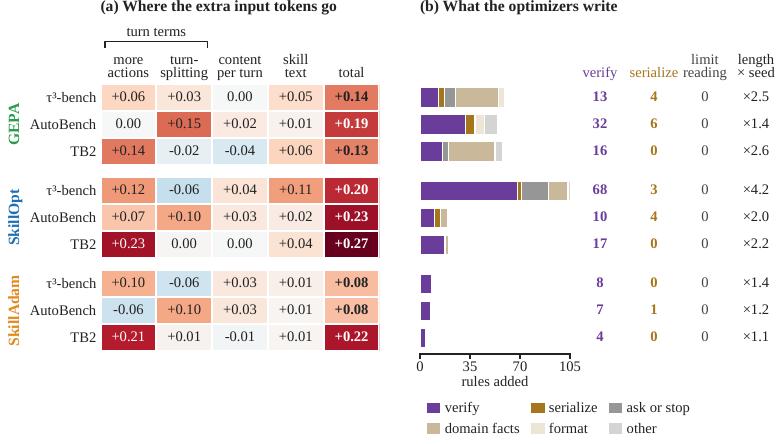}
\caption{\textbf{Over the nine learned skills, extra turns carry most of the extra input tokens, and half of the added rules ask the agent to verify (DeepSeek-V4.1-Flash).} (a) Each skill's change in log input tokens against the seed (total, mean over tasks), split into the four terms of Eq.~\eqref{eq:split}, which add up to it (rounded): more tool calls (more actions), the same calls spread over more turns (turn-splitting), more text per turn (content per turn) and a longer skill (skill text); the first two, the turn terms, sum to the cost of extra turns; red: more tokens than the seed, blue: fewer. (b) Rules added to the seed, by category (Appendix~\ref{app:growth}); verify: re-check a result; serialize: act one call at a time; ask or stop: ask the user, hand off or stop; domain facts: about the domain and its tools; format: how to word the answer; other: the rest, parallelize included. Right: how many added rules verify, serialize or limit reading, and the skill's length as a multiple of the seed's; limit reading is a column only, as none of the nine skills adds such a rule.}
\label{fig:skills}
\end{figure}

Extra turns, more than the longer skill text, carry the extra input tokens. In the four-term split of Eq.~\eqref{eq:split}, on the tasks with six usable episodes (Appendix~\ref{app:components}; Figure~\ref{fig:skills}a), the turn terms carry 73\% of the log-token increase summed over the nine skills and the skill text 21\%, and are larger in 97\% of task-bootstrap draws; the split's first-order remainder is at most 0.001 on average per skill and 0.026 on any task. Under GPT-6-Luna, summed over its six skills on \taubench{} and AutoBench, the turn terms add $+0.86$ to log input tokens and the skill text $+0.04$.

\begin{figure}[t]
\centering
\includegraphics[width=\textwidth]{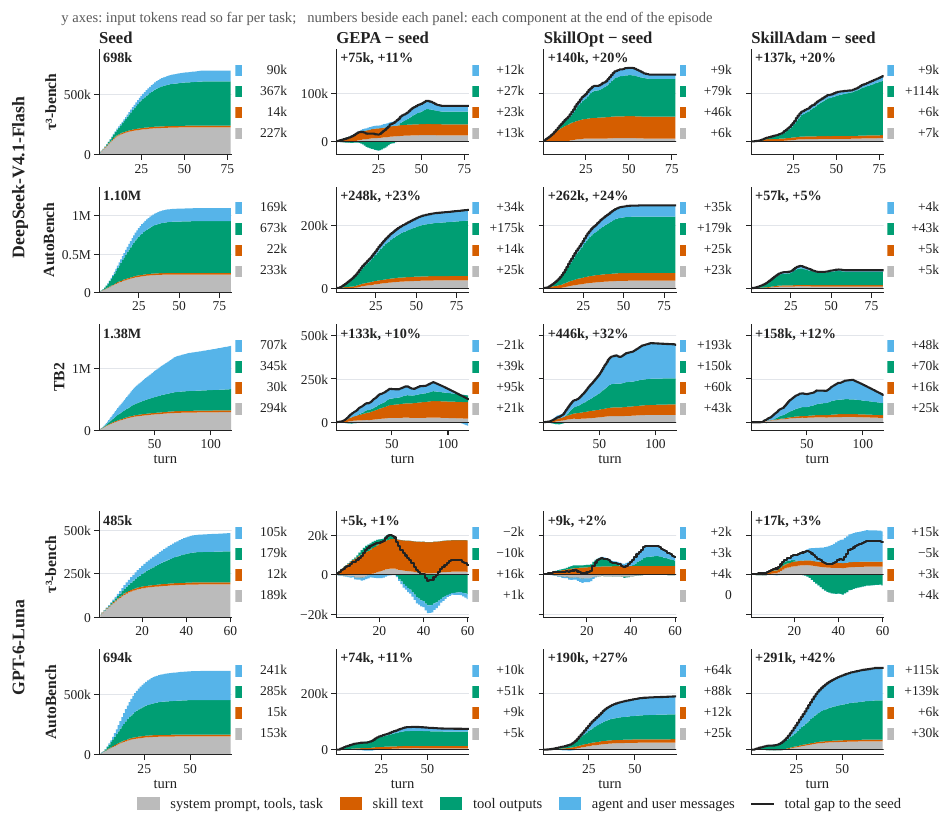}
\caption{\textbf{Tool outputs are the largest part of the extra input tokens for seven of the nine DeepSeek skills and for the three Luna skills on AutoBench.} One row per model and benchmark; curves average over test runs, then over tasks, and an ended episode keeps its total, so a gap can fall where the seed's longer episodes keep reading. Left: the seed, stacked by context component; other columns: learned skill minus seed, black line the total gap, one vertical scale for a row's three gap panels. Bold, top left: the seed's total, or the skill's total gap and that gap as a percentage of the seed's total, a ratio of totals, not $\rho$.}
\label{fig:bipcum}
\end{figure}

Tool outputs are the largest part of the extra input tokens for seven of the nine DeepSeek skills, because the extra turns re-read earlier tool outputs (Figure~\ref{fig:bipcum}); per task the skills read 5 to 32\% more input tokens than the seed, as a ratio of mean totals. SkillOpt on \taubench{} reads 79k more tokens of tool outputs and 46k more of skill text. The exceptions are on TB2: skill text leads for GEPA, 95k of 133k, and agent and user messages for SkillOpt, 193k of 446k. Under Luna, tool outputs lead for the three AutoBench skills, while on \taubench{} the three skills read only 1 to 3\% more than the seed. Figure~\ref{fig:bipcum} draws a row only where every optimizer learned a skill, so it has no Luna TB2 row: there only SkillAdam learned one, which reads 208k (27\%) more than the seed, 195k of it agent and user messages.

\subsection{\texorpdfstring{Cost under Cache Pricing and Compacted History}{Cost under Cache Pricing and Compacted History}}\label{app:mech-cache}\label{app:compaction}

Priced with cached input, which makes up 94 to 98\% of input tokens, every learned skill still costs more than the seed: at 0.1 for a cached and 4 for an output token, relative to an uncached input token, the skills cost 1.05 to 1.27 times the seed, per task as in $\rho$.

Under a compacted history every skill also costs more. Re-billing each recorded episode so that a request pays in full for the prefix but only for the history its last turn added flattens the turn slope from 1.47, the nine skills' mean, to 1.04, as Proposition~\ref{prop:elasticity} predicts, yet $\rho$ is 1.05 to 1.36, every task-bootstrap interval above 1.

\section{\texorpdfstring{Skill Edits}{Skill Edits}}\label{app:rq3}\label{app:skilltext}

\subsection{\texorpdfstring{Skill Growth and Rules}{Skill Growth and Rules}}\label{app:growth}\label{app:coding}

Every optimizer lengthens the skill. Of the 59 DeepSeek-V4.1-Flash edits scored on dev tasks whose parent the logs identify, 58 lengthen the skill; the nine learned skills hold 654 to 2.4k words, against 580 for the seed (Figure~\ref{fig:skills}b).

The added rules ask the agent to verify and to act one call at a time, never to read less. We sort each added rule, labeled by keyword search and checked by hand (a second coder agrees at Cohen's kappa 0.76 on 277 rules), into one of eight categories; four matter here: \emph{verify}, re-check a result against the request; \emph{serialize}, act one call at a time; \emph{parallelize} (under ``other'' in Figure~\ref{fig:skills}b), issue independent calls together; and \emph{limit reading}, read once, read part of an output or keep outputs short. The other four are \emph{ask or stop}, ask the user, hand off or stop; \emph{domain facts}, facts about the domain and its tools; \emph{format}, how to word the answer; and \emph{other}. The nine DeepSeek skills gain 37 rules on average, from 4 to 105, 337 in all: 52\% verify, 18 serialize, one parallelizes reads and none limits reading. Under GPT-6-Luna, 33 of the 38 edits lengthen the skill, and the learned skills add 81 rules: 41\% verify, 3 serialize, two parallelize reads and none limits reading.

\subsection{\texorpdfstring{Tool Calls per Turn}{Tool Calls per Turn}}\label{app:calls}

\begin{figure}[H]
\centering
\includegraphics[width=\textwidth]{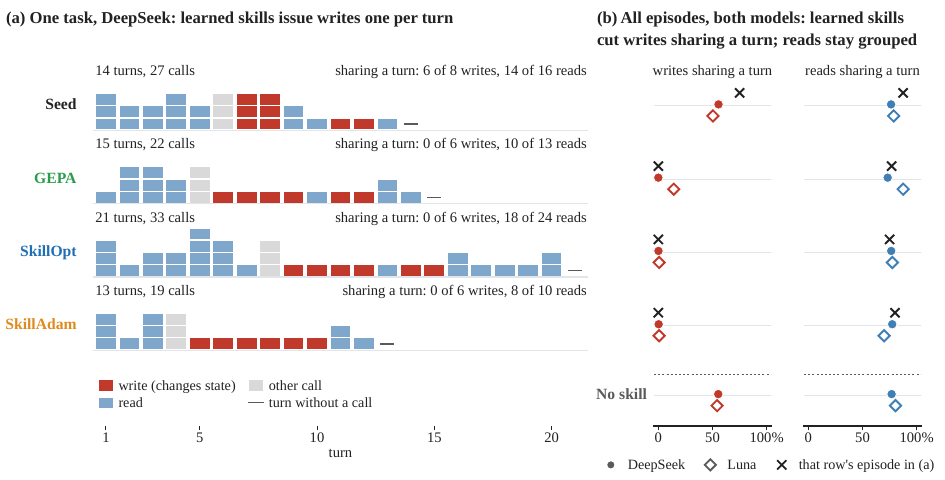}
\caption{\textbf{On AutoBench the learned skills issue writes one per turn while reads stay grouped.} (a) The task of Figure~\ref{fig:teaser}: one column per turn, one square per tool call issued in it; a write (read) shares a turn if another write (read) is issued in the same turn. (b) The same shares, pooled over the writes (reads) of all AutoBench test episodes with complete logs (outcome, tool calls and input tokens all recorded); bottom row: the agent without a skill; a cross marks the share in that row's episode of (a).}
\label{fig:bipwrites}
\end{figure}

Under the learned skills the agent issues its writes one per turn but still groups its reads. On AutoBench a \emph{write} is a call that changes state, and a \emph{read} a GET, a search or a read-only file or shell call, and the rest, such as local file edits, are \emph{other} calls; OpenCode runs all calls of a turn concurrently. Under the seed, 56\% of DeepSeek's writes share a turn with another write; under each learned skill at most 0.3\% do, while the share of reads issued with another read changes by 0.1 to 3.1 points (Figure~\ref{fig:bipwrites}b). Without any skill the agent groups its writes as the seed does, and under Luna the learned skills also cut the share of writes sharing a turn (for GEPA, on the 61 of 120 tasks with complete logs). Because reads stay grouped, merging consecutive single-read turns would save only about one turn per AutoBench episode.

Recomputed with each learned skill's calls packed into turns as densely as the seed packs them, AutoBench's mean $\log\rho$ halves, from 0.17 to 0.08 (Figure~\ref{fig:implications}c), and the bootstrap interval of the remainder excludes zero. The learned skills pack 1.52 to 1.58 calls per turn there, against 1.67 for the seed. On \taubench{} and TB2, where they pack about as many calls per turn as the seed, the recomputation moves the mean $\log\rho$ by at most 0.03.

Most of that halving is serialized writes, which the rules ask for because each AutoBench write loads, changes and saves the whole simulated world, so two concurrent writes can overwrite each other. Packing only each skill's writes into turns as densely as the seed does lowers its log turns by about 0.07 and its $\log\rho$ by about 0.08, nearly all that the recomputation removes.

\subsection{\texorpdfstring{Kept and Rejected Edits}{Kept and Rejected Edits}}\label{app:keptrejected}\label{app:edges}

Selection does not filter cost: kept and rejected edits raise input tokens alike. Of the 59 edits, 30 were kept and 29 rejected; on the dev tasks a candidate and its parent both ran, they raise log input tokens by 0.100 and 0.128 on average, about 10 and 14\%, and a Mann--Whitney rank test finds no difference, $p=0.46$ (Figure~\ref{fig:implications}d).

A gate that also checked cost would have almost nothing to keep (Figure~\ref{fig:implications}d): 44 of the 59 edits raise input tokens over their parent, 73\% of the kept and 76\% of the rejected, and 10 of the other 15 lose dev success, so only 5 keep dev success without adding tokens.

What makes an edit costly is its change in turns, not its length (Figure~\ref{fig:implications}b): over the 59 edits, the change in log turns explains 93\% of the variance of the change in log input tokens, the change in log tool calls 79\% and the change in log skill length 10\%.

\section{\texorpdfstring{Transfer and Trajectories}{Transfer and Trajectories}}\label{app:rq4}

\subsection{\texorpdfstring{Skill Transfer}{Skill Transfer}}\label{app:ood}

\begin{figure}[H]
\centering
\includegraphics[width=6.3in]{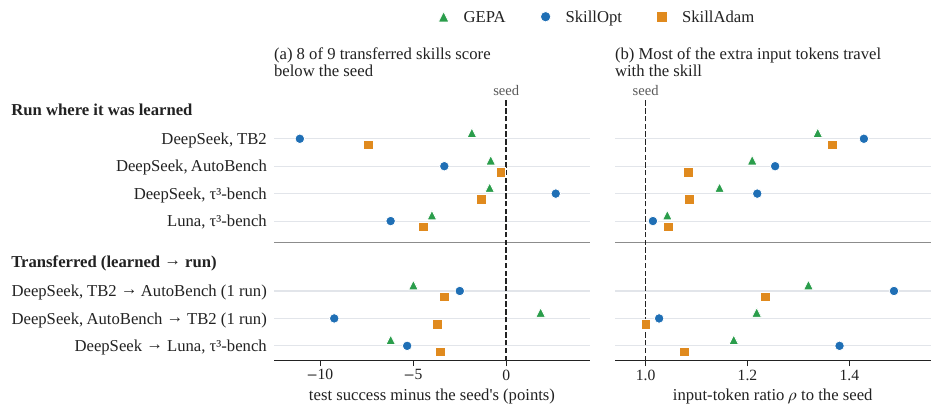}
\caption{\textbf{Transferred skills mostly score below the seed, as they do in the settings they leave or reach, and keep most of their extra input tokens.} Row: the model and benchmark a skill runs on; lower block: learned $\rightarrow$ run, the unchanged part written once; (1 run): one test run per skill, three in all other rows; marker: one optimizer's skill; dashed line: the seed in the same setting.}
\label{fig:transfer}
\end{figure}

The skills do not transfer, and most of their extra input-token cost travels with them (Figure~\ref{fig:transfer}); the upper rows are the four settings these transfers leave or reach, so Luna's TB2 and AutoBench skills, which no transfer uses, are not drawn (Table~\ref{tab:main} has them). Under GPT-6-Luna, three test runs each, the \taubench{} skills learned under DeepSeek read 1.08 to 1.38 times the seed's input tokens and those learned under Luna 1.01 to 1.05. On the other benchmark, one run per skill against a seed run made on the same days, TB2's skills read 1.24 to 1.49 times the seed on AutoBench and AutoBench's 1.00 to 1.22 on TB2.

\subsection{\texorpdfstring{Trajectory Analysis}{Trajectory Analysis}}\label{app:dynamics}

\begin{figure}[t]
\centering
\includegraphics[width=\textwidth]{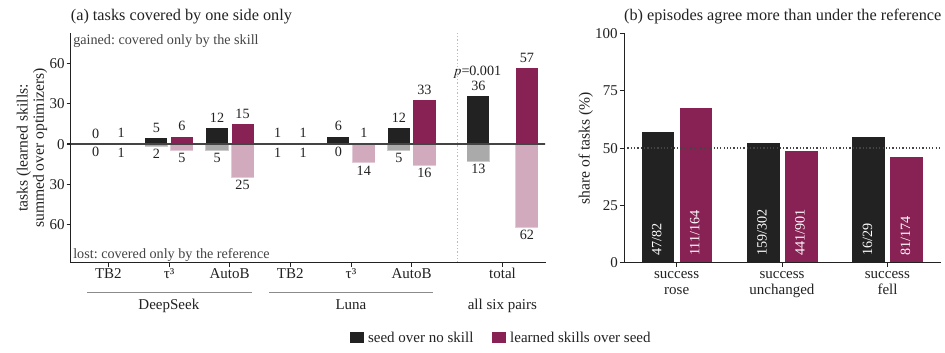}
\caption{\textbf{The seed covers new tasks; the learned skills make episodes agree more on tasks whose success rose.} (a) Tasks covered (solved in at least one of three test runs) by one side only, learned skills summed over optimizers; $p$: sign test. (b) Share of tasks whose episodes agree more than under the reference, by change in success; in bars: such tasks / all tasks; dotted line: half the tasks.}
\label{fig:gains}
\end{figure}

The learned skills sharpen the agent but do not expand it (Figure~\ref{fig:gains}). Over six model--benchmark pairs, the seed gains 36 tasks over the no-skill agent and loses 13 (sign test, $p=0.001$); the learned skills gain 57 over the seed and lose 62, summed over the optimizers (on Luna's TB2 SkillAdam's alone, since GEPA and SkillOpt returned the seed). Where they gain, a task's episodes agree more: for the learned skills against the seed (their bars in Figure~\ref{fig:gains}b), the odds that a task's episodes agree more are 1.61 times as high when its success rose as when it was unchanged (Mantel--Haenszel odds ratio, stratified by the reference's number of successful episodes; $p = 0.03$). The reference is the seed for a learned skill and the no-skill agent for the seed; episodes agree more if their mean pairwise similarity, one minus the Levenshtein distance between two lists of called tool names over the longer one's length, exceeds the reference's.

\subsection{\texorpdfstring{Skill Loading}{Skill Loading}}\label{app:delivery}

\begin{figure}[H]
\centering
\includegraphics[width=\linewidth]{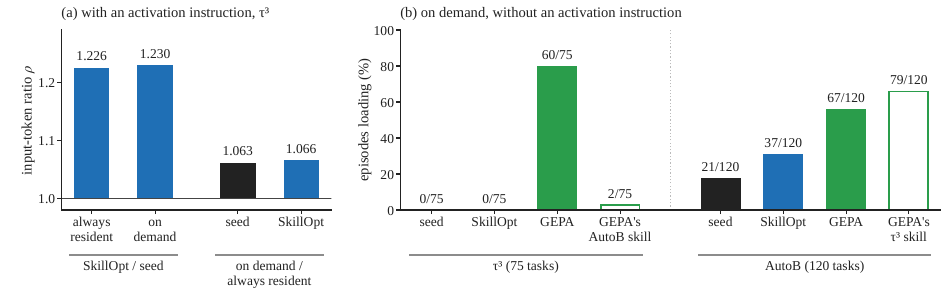}
\caption{\textbf{The route does not change a loaded skill's cost; without an instruction, the description largely decides loading.} DeepSeek-V4.1-Flash, updated release. (a) $\rho$ of SkillOpt's skill against the seed by route (left), and of each skill on demand against always resident (right). (b) Share of episodes that load the skill; label: loaded/total. Hollow: a skill learned on the other benchmark.}
\label{fig:trigger}
\end{figure}

With an activation instruction, the loading route does not change a skill's cost: SkillOpt's \taubench{} skill reads 1.226 times the seed's input tokens always resident and 1.230 on demand (Figure~\ref{fig:trigger}a), and on demand costs both skills 6 to 7\% more, for the skill-call turn. \emph{Always resident}, as in all other runs, the skill follows the task in the first user message; \emph{on demand}, the agent sees only its name and description and loads the body through a skill tool, unprompted or when an \emph{activation instruction} after the task says to. One test run per condition, all on the updated release of DeepSeek-V4.1-Flash (Appendix~\ref{app:harness}).

Without the instruction, the description largely decides whether a skill acts (Figure~\ref{fig:trigger}b). The agent never loads the seed or SkillOpt's skill on \taubench{} (0 of 75); SkillOpt and SkillAdam keep the seed's description in all eleven of their skills. GEPA rewrites it: its \taubench{} skill loads in 60 of 75 \taubench{} episodes, its AutoBench skill in 2 of 75. On AutoBench the agent also loads skills with the seed's description, the seed in 21 of 120 episodes and SkillOpt's skill in 37, and GEPA's AutoBench and \taubench{} skills in 67 and 79.

\section{\texorpdfstring{Skills and Prompts}{Skills and Prompts}}\label{app:skills}

\subsection{\texorpdfstring{Text We Supply to the Optimizers}{Text We Supply to the Optimizers}}\label{app:baselineprompts}

{All three optimizers run their upstream prompt templates unchanged. SkillOpt needs no text from us. We supply the objective that GEPA's generic \texttt{optimize\_anything} template takes (box 1; not the prompt of gskill, its skill optimizer for software tasks) and SkillAdam's five short prompt sections (box 2, with long files cut after their first lines; \texttt{[...]} marks omitted text). \texttt{\{benchmark\}} takes the benchmark's name.}

\begin{promptbox}{1. GEPA: objective added to the upstream template}
\scriptsize
\begin{verbatim}
Improve this agent skill: a Markdown procedure file given to an LLM agent harness as guidance while
  it solves {benchmark} tasks. A better skill lets the agent pass more tasks, as judged by each
  task's own verifier, on unseen tasks of the same benchmark.
If any older skills are still helpful, include them as well. However, always keep the overall
  skills concise and comprehensive.
Adding skills blindly will not help the agent, so only add skills that are actually helpful.
\end{verbatim}
\end{promptbox}

\begin{promptbox}{2. SkillAdam: prompt resource files}
\scriptsize
\begin{verbatim}
[Stand-in for the agent's system prompt]
You are an expert agent solving {benchmark} tasks with the tools your agent harness provides.

[Metric]
The **hard** metric is binary: 1 if the benchmark's verifier judged the task completed, 0 otherwise.
There is no partial credit. A gain of +0.05 means 5% more episodes succeed.

[Success types]
- **success**: task completed successfully

[Failure types]
- **misread_task**: the agent misunderstood, skipped, or invented a task requirement
- **wrong_tool_use**: the agent called the wrong tool or a tool with wrong arguments
- **wrong_sequence**: the agent performed steps in the wrong order
[... 7 more failure types ...]

[Iteration example]
The domain and content below are illustrative only -- do not transfer any
of it into your output.
[... an example JSON reply with "reasoning" and "patch" ...]
\end{verbatim}
\end{promptbox}

\subsection{\texorpdfstring{The Seed Skill}{The Seed Skill}}\label{app:i2}\label{app:seed}

{All runs except those of Appendix~\ref{app:scratch} start from the seed skill \emph{verification-before-completion} \citep{superpowers}, whose rule, run a fresh and complete check before any claim of completion, applies to any task, so one text serves every optimizer and benchmark. The listing is in ASCII: \texttt{[ok]} and \texttt{[x]} stand for check marks and crosses, \texttt{->} for arrows and \texttt{!=} for the not-equal sign.}
\par\smallskip

\lstdefinestyle{seedskill}{basicstyle=\ttfamily\fontsize{6.6}{7.6}\selectfont,breaklines=true,breakindent=1em,columns=fullflexible,keepspaces=true,showstringspaces=false,upquote=true,aboveskip=0pt,belowskip=0pt,frame=none,xleftmargin=0pt}
\begin{tcolorbox}[enhanced,breakable,colback=blue!5!white,colframe=blue!75!black,coltitle=white,fonttitle=\bfseries,title={Seed skill (full text)},arc=2mm,boxrule=1pt,left=1.5mm,right=1.5mm,top=1mm,bottom=1mm]
\begin{lstlisting}[style=seedskill]
---
name: verification-before-completion
description: Use when about to claim work is complete, fixed, or passing, before committing or creating PRs - requires running verification commands and confirming output before making any success claims; evidence before assertions always
---

# Verification Before Completion

## Overview

**Core principle:** Evidence before claims, always.

**Violating the letter of this rule is violating the spirit of this rule.**

## The Iron Law

```
NO COMPLETION CLAIMS WITHOUT FRESH VERIFICATION EVIDENCE
```

If you haven't run the verification command in this message, you cannot claim it passes.

## The Gate Function

```
BEFORE claiming any status or expressing satisfaction:

1. IDENTIFY: What command proves this claim?
2. RUN: Execute the FULL command (fresh, complete)
3. READ: Full output, check exit code, count failures
4. VERIFY: Does output confirm the claim?
   - If NO: State actual status with evidence
   - If YES: State claim WITH evidence
5. ONLY THEN: Make the claim

Skip any step = lying, not verifying
```

## Common Failures

| Claim | Requires | Not Sufficient |
|-------|----------|----------------|
| Tests pass | Test command output: 0 failures | Previous run, "should pass" |
| Linter clean | Linter output: 0 errors | Partial check, extrapolation |
| Build succeeds | Build command: exit 0 | Linter passing, logs look good |
| Bug fixed | Test original symptom: passes | Code changed, assumed fixed |
| Regression test works | Red-green cycle verified | Test passes once |
| Agent completed | VCS diff shows changes | Agent reports "success" |
| Requirements met | Line-by-line checklist | Tests passing |

## Red Flags - STOP

- Using "should", "probably", "seems to"
- Expressing satisfaction before verification ("Great!", "Perfect!", "Done!", etc.)
- About to commit/push/PR without verification
- Trusting agent success reports
- Relying on partial verification
- Thinking "just this once"
- Tired and wanting work over
- **ANY wording implying success without having run verification**

## Rationalization Prevention

| Excuse | Reality |
|--------|---------|
| "Should work now" | RUN the verification |
| "I'm confident" | Confidence != evidence |
| "Just this once" | No exceptions |
| "Linter passed" | Linter != compiler |
| "Agent said success" | Verify independently |
| "I'm tired" | Exhaustion != excuse |
| "Partial check is enough" | Partial proves nothing |
| "Different words so rule doesn't apply" | Spirit over letter |

## Key Patterns

**Tests:**
```
[ok] [Run test command] [See: 34/34 pass] "All tests pass"
[x] "Should pass now" / "Looks correct"
```

**Regression tests (TDD Red-Green):**
```
[ok] Write -> Run (pass) -> Revert fix -> Run (MUST FAIL) -> Restore -> Run (pass)
[x] "I've written a regression test" (without red-green verification)
```

**Build:**
```
[ok] [Run build] [See: exit 0] "Build passes"
[x] "Linter passed" (linter doesn't check compilation)
```

**Requirements:**
```
[ok] Re-read plan -> Create checklist -> Verify each -> Report gaps or completion
[x] "Tests pass, phase complete"
```

**Agent delegation:**
```
[ok] Agent reports success -> Check VCS diff -> Verify changes -> Report actual state
[x] Trust agent report
```

## When To Apply

**ALWAYS before:**
- ANY variation of success/completion claims
- ANY expression of satisfaction
- ANY positive statement about work state
- Committing, PR creation, task completion
- Moving to next task
- Delegating to agents

**Rule applies to:**
- Exact phrases
- Paraphrases and synonyms
- Implications of success
- ANY communication suggesting completion/correctness
\end{lstlisting}
\end{tcolorbox}

\lstdefinelanguage{diff}{morecomment=[f][\color{red!60!black}]-,morecomment=[f][\color{green!40!black}]+,morecomment=[f][\color{blue!60!black}]@@,morecomment=[f][\color{gray}]\ ,breaklines=true,basicstyle=\rmfamily\scriptsize,columns=fullflexible,keepspaces=true,frame=single,xleftmargin=0pt,showstringspaces=false}

\subsection{\texorpdfstring{Skills Learned under DeepSeek}{Skills Learned under DeepSeek}}\label{app:returned-skills}

The nine skills learned under DeepSeek are released in the \texttt{skills/} directory of {\url{https://github.com/valleysprings/BiPA.git}}. All nine optimizer runs end with a skill other than the seed. The shortest is shown below as a line diff against the seed (Appendix~\ref{app:seed}): a line starting with $+$ is added by the skill, and a line starting with $-$ is a seed line it removes.

\paragraph{SkillAdam on TB2.} +4 / -2 lines (654 words vs 580 in the seed).
\begin{lstlisting}[language=diff]
--- seed
+++ SkillAdam
@@ -55,4 +55,5 @@
 - Trusting agent success reports
 - Relying on partial verification
+- Passing only the cases you chose when the acceptance check is external or may be hidden
 - Thinking "just this once"
 - Tired and wanting work over
@@ -70,4 +71,5 @@
 | "I'm tired" | Exhaustion != excuse |
 | "Partial check is enough" | Partial proves nothing |
+| "It passes my tests" | Your cases != the grader's; re-derive every stated requirement literally and try inputs designed to break your solution |
 | "Different words so rule doesn't apply" | Spirit over letter |
 
@@ -94,6 +96,6 @@
 **Requirements:**
 ```
-[ok] Re-read plan -> Create checklist -> Verify each -> Report gaps or completion
-[x] "Tests pass, phase complete"
+[ok] Re-read the task -> list every explicit requirement (exact file name, format/columns/headers, tool + flags, numeric bounds) -> confirm EACH literally on the final artifact -> then try the strictest and most malformed inputs -> fix anything that slips
+[x] "Tests pass, phase complete" / "It works on my cases"
 ```
 
\end{lstlisting}

\lstdefinestyle{scratchskill}{breaklines=true,basicstyle=\rmfamily\scriptsize,columns=fullflexible,keepspaces=true,frame=single,xleftmargin=0pt,showstringspaces=false}

\subsection{\texorpdfstring{Skills Learned from the Three-Word Seed}{Skills Learned from the Three-Word Seed}}\label{app:scratch-skills}

The GPT-6-Luna runs of Appendix~\ref{app:scratch} start from the three-word seed below; the four skills they return follow verbatim. Word counts include the front matter.

\paragraph{Three-word seed.} 13 words.
\begin{lstlisting}[style=scratchskill]
---
name: minimal
description: General guidance for solving tasks
---

Solve the task.
\end{lstlisting}
\paragraph{SkillOpt, AutoBench.} 347 words.
\begin{lstlisting}[style=scratchskill]
---
name: minimal
description: General guidance for solving tasks
---

For tool-driven tasks, initialize the available tools and inspect their usage guidance, service limits, and supported operations before acting. Respect applicable privacy, authorization, availability, hold, and status constraints. Make only necessary, in-scope changes. If verification finds a missing or inconsistent change, retry only what is needed and verify again. Report completed actions clearly and distinguish them from requests or actions still pending.

## Workflow Automation
Before acting, identify the authoritative source records, applicable policies and exceptions, eligibility conditions, required actions, and exact destination for each result. Follow explicit instructions and structured policies as written; do not invent eligibility rules from incidental notes or infer that approval covers actions it does not. Complete every independently required communication, record update, and follow-up action for eligible cases, while respecting skip rules and exclusions.

Use the supported operation and exact resource or relationship that matches each requested outcome; a nearby substitute, such as a related note instead of a description field, does not satisfy the requirement. Match the destination schema precisely, including field names, column or label IDs, value types, and documented request structure. Pass query parameters and body fields through the documented tool arguments. Preserve source values verbatim when required.

Avoid overlapping state-changing calls to shared services or records; perform mutations sequentially, especially for batches. A successful response alone is not proof that a change persisted. Read back affected records or use an appropriate verification operation, confirm the intended entity, recipient or channel, fields, values, and relationships, and correct discrepancies before proceeding or claiming completion.

Before sending a report or alert, check it against the requested fields and relevant source records. Verify that each message is recorded for the intended recipient and explicitly includes required names, identifiers, values, counts, labels, dates, reference strings, and context; do not rely on readers to infer details from a list. If a task requires a particular service or artifact, attempt and verify that action. If the required service is unavailable, state the limitation accurately rather than claiming completion or presenting a substitute as equivalent.
\end{lstlisting}
\paragraph{GEPA, AutoBench.} 388 words.
\begin{lstlisting}[style=scratchskill]
---
name: workflow-automation
description: Reliable, policy-aware procedures for completing automation tasks
---

Complete the requested workflow using the available tools. Do not ask clarifying questions. First run `python /workspace/automationbench_cli.py init`, read `/workspace/tooling.md`, and check `/workspace/agent_config.json` for allowed services and initial task data.

## Inspect and Decide

- Discover relevant API endpoints with `api_search`; use the exact endpoint and request schema. In API mode, pass query parameters through `--params` as JSON rather than appending them to the URL.
- Read the source records and applicable procedures before acting. Check for current policy updates and relevant communications from authorized senders; apply effective dates and explicit exceptions. Prefer current authoritative policy over older guidance, incidental requests, and unrelated or untrusted content.
- Determine the exact in-scope records and required actions before making changes. Respect date windows, statuses, opt-ins, approval requirements, exclusions, and scope limits. Do not invent violations or requirements beyond the written policy.
- Preserve source values verbatim in notifications and records. Include all requested names, identifiers, companies, references, amounts, and counts. When totals are requested, calculate them from the eligible source records, show the appropriate grouping, and include the total in the message.

## Execute Safely

- Read the tool guide for the active toolset. Use the requested service or artifact for each deliverable; do not substitute a narrative summary for a requested record, document, or generated content.
- Perform independent reads in parallel when useful. Perform mutations sequentially, especially when they update the same service or persisted world state. Concurrent writes can silently overwrite one another.
- Send API-mode Gmail messages using correctly base64url-encoded RFC 2822 content via `base64_encode`. Use actual line breaks in the message before encoding; do not send literal `\\n` text.
- When creating drafts, ensure they remain drafts if the instructions say not to send. Do not perform consequential actions that policy forbids or that require unavailable information.

## Verify and Report

- After mutations, read back the affected records and confirm every requested change persisted: statuses, tags, tracking rows, destinations, event attendees, document state, or generated artifacts as applicable. A successful write response alone is not sufficient.
- If verification reveals a failed or overwritten update, correct it with a sequential write and verify again.
- Summarize only completed actions and the items acted on. Do not narrate skipped, excluded, or rejected items.
\end{lstlisting}
\paragraph{SkillOpt, \taubench{}.} 330 words.
\begin{lstlisting}[style=scratchskill]
---
name: minimal
description: General guidance for solving tasks
---

Solve the task.

### Track and Complete Requested Actions
For tasks involving multiple items or follow-up actions, keep a checklist of each requested target and its required next step. Inspect all relevant records rather than stopping at the first match, and handle each eligible target separately. When a documented tool is callable, use it for authorized actions and follow its documented workflow and arguments exactly. For user-discoverable tools, distinguish providing or unlocking the tool from actually executing it; do not invent tools or use agent-only tools on the user's behalf. In troubleshooting, work through applicable in-scope checks and fixes, verify the results, and escalate only when those steps are exhausted or a documented rule requires escalation. Do not report an action as complete without a confirming result. Before ending, compare the checklist with the results and identify anything blocked or incomplete.

### Keep the Conversation Open
After each user response, address all outstanding requests in that response before ending the conversation. Do not end while a question or requested action remains unanswered; end only when the case is resolved, appropriately transferred, or the user clearly asks to stop.

### Verify Calculations
Use an available calculator tool for derived amounts such as totals, price differences, refunds, and remaining balances. Base calculations on verified source values, check the result against the intended action, and report the verified amount accurately.

Match each requested item to its exact record and keep actions within the scope the user authorized. Preserve unrelated items or settings, and use a broader operation only when it matches the request. Clearly distinguish completed, unavailable, and unresolved parts. If an available operation would affect additional items or records, do not proceed; explain the limitation.

For guided troubleshooting, give one actionable step at a time and wait for the user's result before choosing the next step. Use that result to guide further checks, and retest the original issue after making a fix.
\end{lstlisting}
\paragraph{GEPA, \taubench{}.} 438 words.
\begin{lstlisting}[style=scratchskill]
---
name: tau3-task-resolution
description: Policy-first workflow for completing tau3-bench customer-service tasks with domain tools
---

## Operating Procedure

1. Call `start_conversation` exactly once. Follow the task's instructions and policy as the source of truth; do not assume one domain's rules apply to another.
2. Work in small sequential steps. Make at most one tool call per step, and do not combine a tool call with a user message.
3. Identify the user before accessing their records, using the task's required lookup and authentication process. Complete any required verification logging before proceeding.
4. Inspect relevant records and authoritative knowledge-base guidance before recommending or changing anything. Use the exact tool names, arguments, and order specified there. Do not invent facts, prices, eligibility, policies, or tool capabilities.
5. Resolve the user's actual request, not just a nearby issue. If they ask for a review of all relevant records, inspect all relevant records and categories, not only the first examples. Calculate from documented rates and rules; distinguish missing, correct, and excess rewards rather than treating every surprising value as an underpayment.
6. Before a database-changing action, state the exact action and material consequences, then obtain explicit confirmation when policy requires it. For one-time actions, collect every item and detail first. After acting, inspect the result and confirm the final state.
7. Distinguish tools the agent must use from tools intended for the user. Follow the knowledge base's instructions for discoverable tools exactly: unlock and call agent tools only when directed; give user tools only when directed. If a user must execute a discoverable tool, provide the exact required identifiers and arguments, and help them proceed without claiming the action is complete until confirmed.
8. For technical troubleshooting, follow the documented diagnostic and fix sequence, one step at a time when the user needs guidance. Reassess after each result and continue through all applicable, supported fixes. For MMS, check prerequisites and work through relevant causes such as service/SIM, network mode, mobile data and roaming, APN/MMSC, Wi-Fi Calling, and messaging-app permissions. If APN settings are reset, complete the required reboot before concluding.
9. Do not transfer merely because a step is inconvenient, an answer is uncertain, or the user requests a human while the issue remains within scope and can still be resolved under policy. Try all applicable supported actions first. Transfer when policy requires it or the task cannot be handled, using the prescribed transfer sequence and exact user-facing message.
10. Keep customer-facing replies concise, factual, and reassuring without unsupported assurances. Do not expose internal notes or speculate. When the task is resolved, report the verified outcome and end the conversation as instructed.
\end{lstlisting}

\end{document}